\documentclass[11pt]{article}

\usepackage[preprint]{acl}

\usepackage{times}
\usepackage{latexsym}

\usepackage[T1]{fontenc}

\usepackage[utf8]{inputenc}

\usepackage{microtype}

\usepackage{inconsolata}

\usepackage{graphicx}

\title{ViLL-E: Video LLM Embeddings for Retrieval}

\author{Rohit Gupta \and 
  Jayakrishnan Unnikrishnan \and
  Fan Fei \and
  Sheng Liu \\
  \texttt{\{rohitgpt, jayunn, feiff, shenlu\}@amazon.com} \\\AND
  Son Tran \\
  \texttt{sontran@amazon.com} \\\And
  Mubarak Shah \\
  \texttt{shah@crcv.ucf.edu} 
}

\usepackage[T1]{fontenc}                    

\usepackage{microtype}

\usepackage{amsmath}
\usepackage{amssymb}                        
\usepackage{pifont}
\usepackage{amsfonts}

\usepackage[dvipsnames,table]{xcolor}       
\usepackage{graphicx}
\usepackage{tikz}
  \usetikzlibrary{patterns}
  \pgfdeclarelayer{background}
  \pgfsetlayers{background,main}

\usepackage{booktabs}                       
\usepackage{multirow}
\usepackage{caption}
\usepackage{subcaption}
\usepackage{float}
\usepackage{siunitx}
\usepackage{enumitem}
\usepackage[export]{adjustbox}  
\usepackage{wrapfig}   
\usepackage{pifont}
\usepackage{amsmath}
\usepackage{amsfonts}
\usepackage{geometry}
\usepackage{xcolor}
\usepackage{ifsym}
\usepackage{changepage}                        

\usepackage[most]{tcolorbox}

\newcommand{\ColorBox}[2]{%
  \raisebox{-0.5pt}[0pt][0pt]{\tcbox[
    colframe=#1,
    colback=#1!20!white,
    boxrule=0.3mm,
    arc=0mm,
    boxsep=0pt,
    left=1pt,
    right=1pt,
    top=0.35pt,
    bottom=0.35pt,
    valign=center
  ]{#2}}%
}

\usepackage{fontawesome5}

\definecolor{Set1Red}{HTML}{E41A1C}
\definecolor{Set1Blue}{HTML}{377EB8}
\definecolor{Set1Green}{HTML}{4DAF4A}
\definecolor{Set1Purple}{HTML}{984EA3}
\definecolor{Set1Orange}{HTML}{FF7F00}
\definecolor{Set1Yellow}{HTML}{FFFF33}
\definecolor{Set1Brown}{HTML}{A65628}
\definecolor{Set1Pink}{HTML}{F781BF}
\definecolor{Set1Gray}{HTML}{999999}
\definecolor{readableyellow}{HTML}{B58900} 

\newcommand{\limited}{\textcolor{orange}{\ding{117}}} 

\definecolor{cadmiumgreen}{rgb}{0,0.42,0.24}
\definecolor{maroon}{rgb}{0.5,0,0}
\definecolor{britishracinggreen}{rgb}{0,0.26,0.15}

\newcommand{\Yes}{\ding{51}}

\newcommand{\FT}{{\tiny \texttt{FT}}}
\newcommand{\RetSymb}[1]{\footnotesize{\texttt{R@#1}}}

\begin{document}
\bibpunct{[}{]}{,}{n}{,}{,}
\maketitle

\begin{abstract}

Video Large Language Models (VideoLLMs) excel at video understanding tasks where outputs are textual, such as Video Question Answering and Video Captioning. However, they underperform specialized embedding-based models in Retrieval tasks, such as Text-to-Video Retrieval and Moment Retrieval. We introduce ViLL-E (Video-LLM-Embed), a unified VideoLLM architecture endowed with a novel embedding generation mechanism that allows the model to ``think longer'' for complex videos and stop early for easy ones. We train this model with a three-stage training methodology combining generative and contrastive learning: initial large-scale pre-training with video-caption pairs; followed by continual training on a smaller, detailed-caption dataset; and concluding with task-specific fine-tuning on a novel multi-task dataset covering Video QA, Temporal Localization, Video Retrieval, and Video-Text Matching. Our model significantly improves temporal localization (on avg. $7$\% over other VideoLLMs) and video retrieval (up to 4\% over dual encoder models), achieving performance comparable to state-of-the-art specialized embedding models while remaining competitive on VideoQA tasks. Furthermore, our joint contrastive-generative training unlocks new zero-shot capabilities, significantly outperforming state-of-the-art methods in composed video retrieval (+5\% over SotA) and retrieval from long text (+2\% over SotA).


\end{abstract}
\section{Introduction}
\label{sec:intro}
\vspace{-0.5em}


\newcommand{\cmark}{\textcolor{cadmiumgreen}{\ding{51}}}
\newcommand{\xmark}{\textcolor{red}{\ding{55}}}

\begin{figure*}[t]
  \centering
  \begin{subfigure}[t]{0.63\linewidth}
  \vspace{0pt}      
    \centering
    \includegraphics[width=\linewidth,trim={0.1cm 0 0.2cm 0},clip]{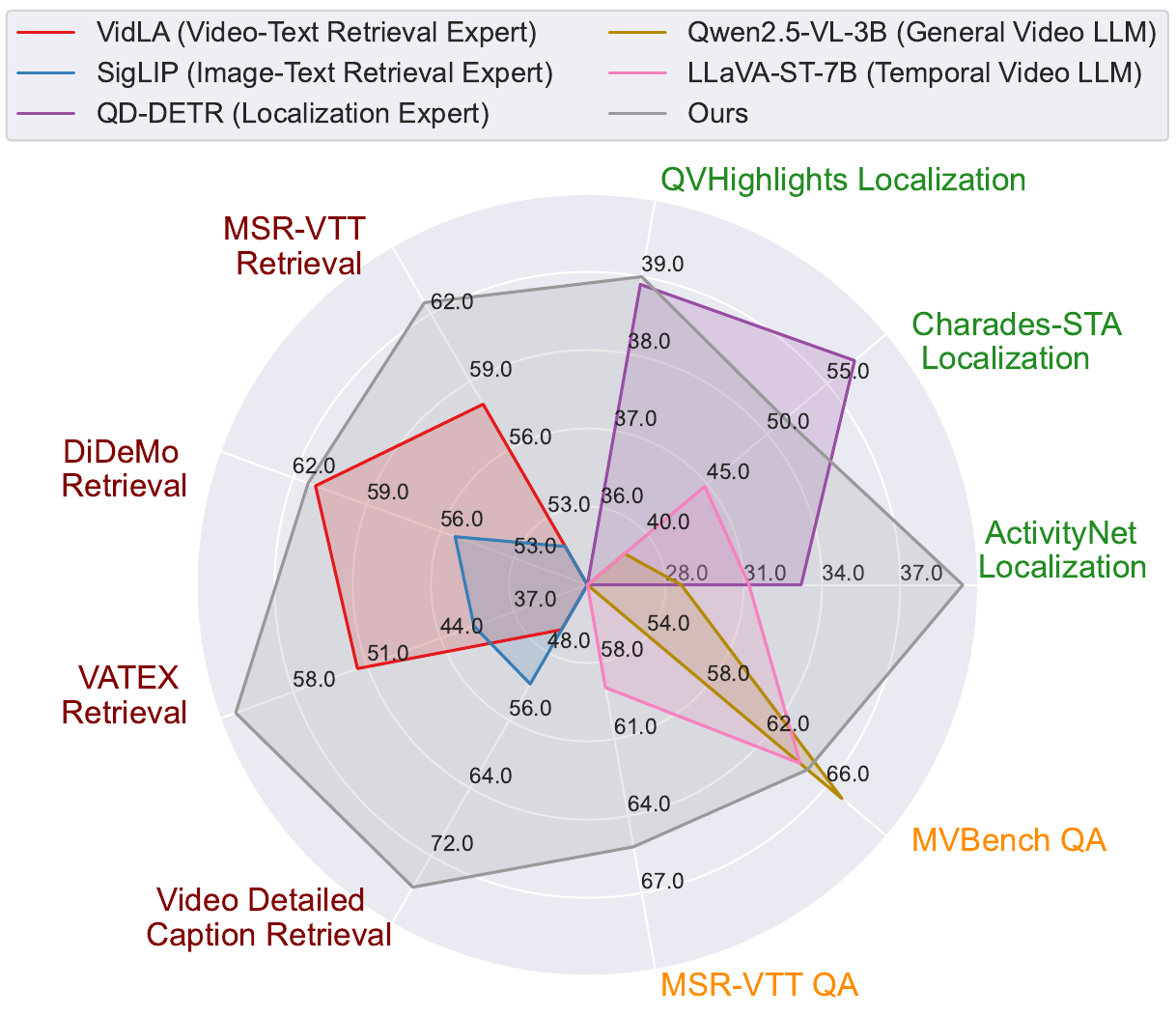}
    \caption*{}
  \end{subfigure}%
  \hfill
\raisebox{-5\baselineskip}{%
\begin{subfigure}[t]{0.36\linewidth}
  \centering
  \small   

  \setlength{\tabcolsep}{2pt}
    \renewcommand{\arraystretch}{1.2}

  \begin{tabular}{@{}lcc@{}}
    \toprule
    & \multicolumn{2}{c}{\textbf{Architecture}} \\
    \textbf{Method} & \textbf{LLM} & \textbf{Embeddings} \\
    \midrule
    \ColorBox{Set1Red}{VidLA}            & \xmark & \cmark \\
    \ColorBox{Set1Blue}{SigLIP}           & \xmark & \cmark \\
    \ColorBox{Set1Purple}{QD‑DETR}          & \xmark & \cmark \\
    \midrule
    \ColorBox{readableyellow}{Qwen2.5‑VL}    & \cmark & \xmark \\
    \ColorBox{Set1Pink}{LLaVA‑ST}      & \cmark & \xmark \\
    \midrule
    \ColorBox{Set1Gray}{Ours}             & \cmark & \cmark \\
    \bottomrule
  \end{tabular}

  \vspace{0.7em}  

  \setlength{\tabcolsep}{1pt}
  \renewcommand{\arraystretch}{1.2}
  \vspace{-0.5em}

  \begin{tabular}{@{}lcccc@{}}
    \toprule
    & \multicolumn{4}{c}{\textbf{Capabilities }} \\
    \textbf{Method} & \textbf{\textcolor{Mahogany}{Ret.}} & \textbf{\textcolor{cadmiumgreen}{Loc.}} & \textcolor{orange}{\textbf{QA}} & \textbf{Cap.} \\
    \midrule
    \ColorBox{Set1Red}{VidLA}            & \cmark & \xmark & \xmark & \xmark \\
    \ColorBox{Set1Blue}{SigLIP}           & \cmark & \xmark & \xmark & \xmark \\
    \ColorBox{Set1Purple}{QD‑DETR}          & \xmark & \cmark & \xmark & \xmark \\
    \midrule
    \ColorBox{readableyellow}{Qwen2.5‑VL}    & \xmark & \limited & \cmark & \cmark \\
    \ColorBox{Set1Pink}{LLaVA‑ST}      & \xmark & \limited & \cmark & \cmark \\
    \midrule
    \ColorBox{Set1Gray}{Ours}             & \cmark & \cmark & \cmark & \cmark \\
    \bottomrule
  \end{tabular}

  \caption*{}   
\end{subfigure}%
}

  \vspace{-2em}
  \caption{{\bf Left \& Bottom Right}: VideoLLMs lag expert models on some retrieval‑based tasks, e.g.\ Temporal Localization (in \textcolor{cadmiumgreen}{green}~\cite{qvhighlights, charadessta, activitynetcap}), and are incapable of others, e.g.\ Text‑to‑Video Retrieval (in \textcolor{Mahogany}{red}~\cite{msrvtt, didemo, vatex}). A key difference between existing VideoLLMs and state of the art expert models in these tasks is the use of embeddings( {\bf Top Right}). Our approach, \textbf{ViLL‑E} (VideoLLM‑Embed, \textit{pronounced willy}) equips VideoLLMs with embedding generation. ViLL-E matches SotA on retrieval tasks while remaining competitive on generation tasks such as VideoQA (in \textcolor{orange}{orange}~\cite{msrvttqa, mvbench}). \limited ~indicates limited capabilities.}
  \label{fig:teaser}
  \vspace{-1.5em}
\end{figure*}


Large Language Models (LLMs) have demonstrated significant capabilities in a ``single model multi-task approach'' setting, effectively solving various natural language understanding tasks. Inspired by the success of LLMs~\cite{gpt3,llama,mistral,t5}, numerous studies have integrated additional modalities like images~\cite{blip2,llava,qwenvl} and audio~\cite{audiopalm,speechgpt} into these models.  These Multimodal LLMs models excel at most vision tasks that can be formulated as text {\em generation problems}. Of specific interest are Video Large Language Models (VideoLLMs)~\cite{videochatgpt,videollava,videollama,videochat}, which incorporate videos into LLMs by encoding video frames and aligning visual features with LLM feature spaces via projection layers, enabling video understanding through tasks like captioning and QA.

However, this generative approach has limitations. Many video tasks, like Text-to-Video Retrieval (T2V), require embedding-based matching rather than text generation. Such tasks typically use specialized models with aligned video-text embeddings, fine-tuned per dataset. Currently these specialized methods lead performance on standard benchmarks. Even VideoLLM variants that are specialized to attempt temporal reasoning (e.g. LLaVA-ST~\cite{llavast}) trail expert detectors such as QD-DETR~\cite{qddetr} by double-digit margins on ActivityNet~\cite{activitynetcap} and Charades-STA~\cite{charadessta}. 
Figure 1 underscores the point: every top performer in video retrieval, temporal localisation or cross-modal search (VidLA~\cite{vidla}, QD-DETR~\cite{qddetr}, SigLIP~\cite{siglip}) achieves its performance with embeddings, while Video-LLMs dominate the Video-QA column. Also, recent NLP research has demonstrated that LLMs can be transformed into strong retrieval models with contrastive finetuning on surprisingly little data (see, e.g., GRIT~\cite{grit} or E5~\cite{e5}). This raises the question: Instead of maintaining two separate stacks, can a single model handle both generative tasks and extract discriminative video/text embeddings, addressing the retrieval and localization gaps?



Motivated by these observations, we propose an approach that combines generative and embedding tasks during Video-LLM training, aiming for strong performance on generative tasks like VideoQA, while also enabling embedding generation for tasks such as video retrieval—previously unsupported by VideoLLMs. Our approach builds on PaliGemma~\cite{paligemma}, a standard multi-modal LLM and adds a learnable attentive-pooling “embedding head” after the LLM decoder.  The embedding head is designed to adapt to video complexity during  inference. This allows the model to ``think longer'' for hard videos but return quickly for easy ones. We propose three-stage training, which begins with large-scale pre-training on video-caption pairs using joint supervision: next-word prediction for generation and a multi-modal contrastive loss for embedding. In the second stage, we continue pre-training on a smaller dataset of videos with high-quality generated captions. The final stage involves instruction finetuning on a multi-task dataset covering VideoQA, captioning, localization, and retrieval. 
Our proposed model, ViLL-E (pronounced willy), is competitive with VideoLLMs on VideoQA and outperforms them on localization. Our embeddings narrow the gap with non-LLM localization expert methods and enable SoTA performance on video retrieval.

\noindent In summary, the key contributions of our work are:
\vspace{-1.6em}
\begin{itemize}[leftmargin=1em,itemsep=0.25em,parsep=0pt]
  \item We present ViLL-E, an unified solution to diverse video-understanding tasks,  which is the first VideoLLM that can generate text responses or video \& text embeddings with a single model.
  \item An effective three-stage training pipeline: (1) large-scale joint captioning + contrastive learning, (2) continued pre-training on high quality captions, and (3) multitask fine-tuning.
  \item Our single model ViLL-E, outperforms VideoLLMs by more than 7\% across three moment retrieval benchmarks, rivals them on VideoQA, and is competitive with expert retrieval models.
  \item ViLL-E also solves new tasks in zero-shot setting, including long-caption retrieval, compositional video search, and two-stage retrieve-and-match pipelines using a single model.
\end{itemize}

\vspace{-1.5em}
\section{Related Work}
\label{sec:related}
\vspace{-0.75em}

\noindent \textbf{Video Expert Models for Retrieval and Localization:} Prior works on text-video and video-text  retrieval~\cite{clipvip,clip4clip,xpool,vidla, internvideo2} typically extend dual-encoder image-text models such as CLIP~\cite{clip} or BLIP~\cite{li2022blip} to video by aggregating the CLIP vision encoder outputs from multiple frames with architecture adaptations to model temporal relations~\cite{timesformer, chen2022litevl}. For retrieval they use dense vector search in the embedding space while a few other methods use an additional reranking step with cross-modality fusion layers~\cite{umt23,vindlu}. Prior works on temporal localization such as Moment-DETR~\cite{qvhighlights}, QD-DETR~\cite{qddetr}, TR-DETR~\cite{trdetr} and UnLoc~\cite{unloc} build on ideas from retrieval. Typically Video-Text Fusion is also employed to score the match between a given text and video segment for compatibility. Specific unique ideas include additional modules like time span prediction in QD-DETR and learnable moment queries in Moment-DETR. Vid2Seq~\cite{vid2seq} utilizes dense captioning trained on large video-text datasets instead of retrieval for localization. These expert models are close to the state of the art frontier for their respective tasks, however they are specialized for one task, and are  typically finetuned individually for each dataset.


\noindent \textbf{Video LLMs:} Many recent works have extended ImageLLMs such as  LLaVA~\cite{llava} to videos. VideoLLaVA~\cite{videollava} is a straightforward extension that utilizes LanguageBind~\cite{languagebind} encoder instead of CLIP because its trained on both images and videos. Some works have tried to extend a BLIP2 style architecture to Videos, namely VideoChat~\cite{videochat} and Video LLaMA~\cite{videollama}. In addition, some works have focused on creation of strong IFT datasets specific to video tasks, VideoChatGPT~\cite{videochatgpt} created IFT data using ActivityNet dense captions and semi-automated process with questions generated by GPT3.5 using dense text captions and open vocabulary attributes. In VideoChat2~\cite{videochat}, another  diverse IFT dataset was created  by combining many (30+) datasets together. These models excel at video question answering and other text generation tasks, however they often struggle at temporal localization tasks and don't generate embeddings, which are necessary for video retrieval.


\noindent \textbf{Video LLMs with Localization IFT:} Recent works have sought to instill temporal awareness and localization skills in video LLMs. Broadly, the architectures of these approaches can be categorized into LLAVA-like or BLIP2-like~\cite{blip2} depending on whether or not they use a Q-Former for token reduction before input to LLM. TimeChat~\cite{timechat} uses sliding window Q-Formers as a projection layer, VTimeLLM~\cite{vtimellm} uses MLP, HawkEye~\cite{hawkeye} uses simple Q-Former and Momentor~\cite{momentor} proposes a parallel temporal perception module as part of its projection layers. LLaVA-ST~\cite{llavast} inserts special spatio-temporal tokens into the language stream, and aligns them with vision features through a language-aligned position embedding. To impart temporal awareness to the models, these works use IFT datasets with specialized tasks such as dense captioning, timestamp generation for localization, repetition counting and more. On localization tasks, these models outperform general VideoLLMs, however they are unable to outperform expert models which do not use LLMs.


\noindent \textbf{LLMs for Embedding generation:} Some works in the NLP domain~\cite{e5, nvembed} have demonstrated that generative LLMs can be finetuned with a limited amount of data (relative to the scale of generative pre-training) to generate strong embeddings for a variety of downstream tasks including retrieval. GRIT~\cite{grit} has demonstrated that a pre-trained LLM can be jointly finetuned to improve on both generative instruction following and embedding generation, maintaining strong performance on both sets of tasks. Some recent concurrent works such as VLM2Vec~\cite{vlm2vec}, GME~\cite{gme} and MM-Embed~\cite{mmembed} have sought to use LLMs for multi-modal embeddings, but are limited to images. 

Our approach unifies text-generative and embedding-generative capabilities into a single VideoLLM model, which is competitive with task-specific and dataset-specific expert models on temporal localization and retrieval tasks that require embedding generation, as well as with Video LLMs on text generation like QA and captioning tasks.


\begin{figure*}[t]
    \centering
    \includegraphics[width=0.9\linewidth]{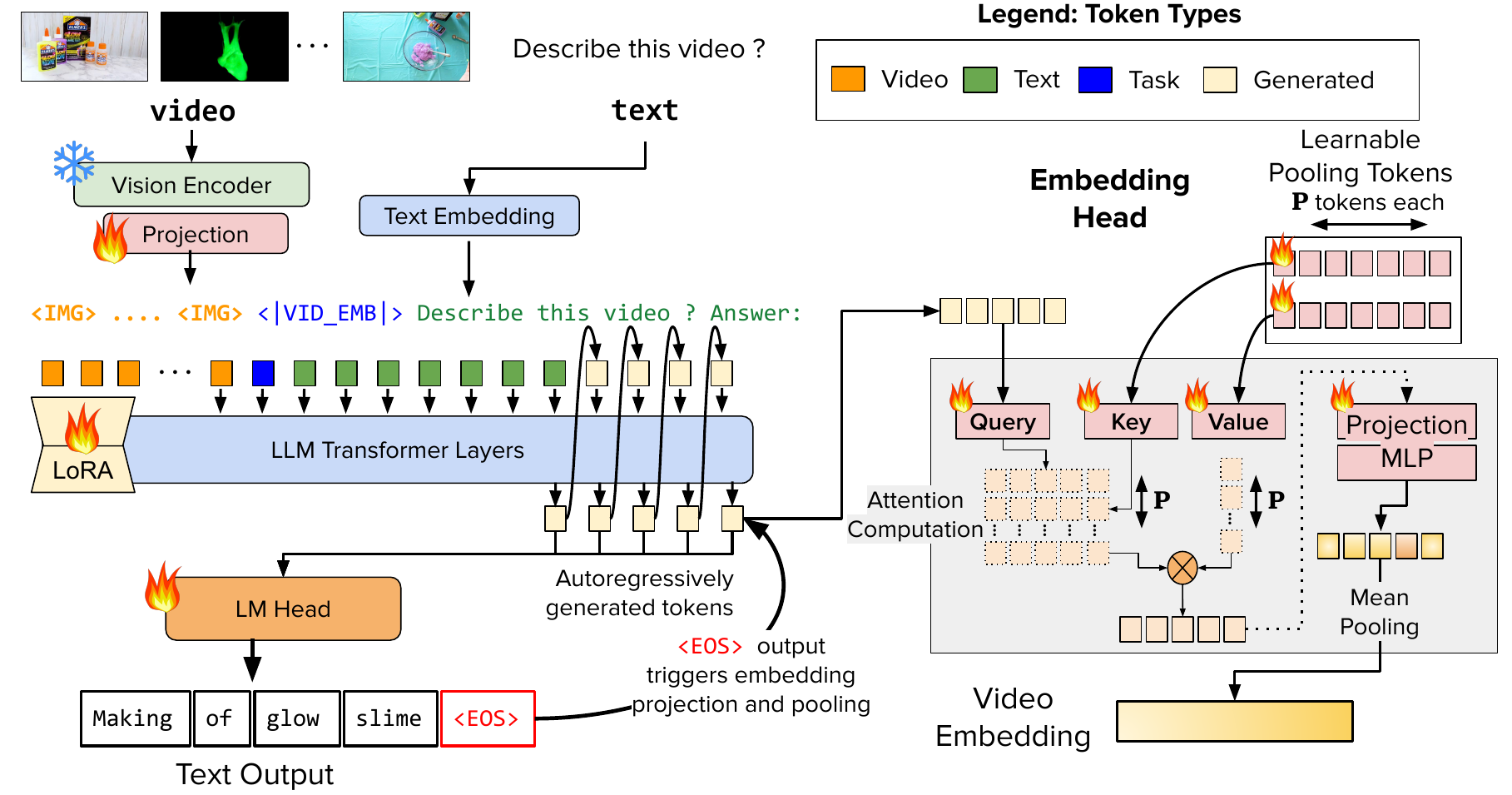}
    \vspace{-0.25em}
    \caption{Our model ViLL-E is a multi-modal LLM which has been equipped with  additional embedding generation capabilities. Individual video frames are first encoded by a pretrained vision encoder, with extracted features projected into embedding vectors. These visual embeddings are combined with textual embeddings of an input prompt and jointly processed autoregressively through a pre-trained Multi-Modal LLM's transformer layers. After the model autoregressively generates an end-of-sequence (\texttt{|EOS|}) token, all the generated tokens are collected and passed to our embedding head. The embedding head employs an attention-based learnable pooling mechanism to extract and  aggregate relevant information across the tokens; The generated tokens are further passed through projection layers and mean pooled to yield a compact, informative video embedding. The ability to dynamically generate different number of tokens to feed the embedding head allows the model to handle videos of different complexity levels and is key for its success at varied retrieval tasks.}
    \label{fig:system}
    \vspace{-1.25em}
\end{figure*}

\section{Method}
\label{sec:method}
\vspace{-0.5em}

In this section, we present the architecture and training methodology of our VideoLLM, designed to leverage multi-modal video data to perform both generative and embedding tasks. Our approach involves a multi-stage and multi-task process that includes continued multi-modal pre-training to improve video language alignment, and subsequent multi-task finetuning.


\vspace{-0.5em}
\subsection{Model Architecture}
\vspace{-0.5em}

Our model (see Figure~\ref{fig:system}) features an LLM backbone, shared by both its generative and embedding tasks, integrated with a vision encoder module designed to effectively process video inputs. We initialize the model's weights using a pretrained multimodal LLM (PaliGemma-3B~\cite{paligemma}). Following PrefixLM (Raffel et al., 2020)~\cite{t5}, visual tokens and input prompts are processed with bidirectional attention, whereas causal attention is applied to the suffix autoregressively generated by the model. For text generation tasks (e.g. VideoQA), our model follows the same approach as PaliGemma.

\subsection{Embedding Head Design}
\label{sec:embedhead}

We introduce a learnable embedding head to support retrieval-oriented tasks with four design goals: (1) dedicated modeling capacity separate from generative tasks, (2) a bottleneck that yields dense embeddings without overly restricting representation power, (3) adaptability to variable token lengths for videos of different complexity, and (4) parameter efficiency. We evaluated several approaches:  
\textbf{1. Attention-Free pooling}: Pools penultimate layer LLM tokens into a fixed-size embedding, lightweight but dependent on heuristic pooling (cf. Sentence-BERT~\cite{reimers-2019-sentence-bert}).  
\textbf{2. Self-Attention}: Adds an extra transformer layer, offering some task-specific capacity but redundant with prior LLM layers and lacking a bottleneck.  
\textbf{3. Q-Former}: Learned queries cross-attend to LLM outputs to produce fixed-length embeddings (e.g., BLIP-2~\cite{blip2}); effective but inflexible for variable content length, as output size always equals number of learnable queries.  
\textbf{4. `K-Former'}: Learns only keys, enabling variable-length outputs.  
\textbf{5. Our `KV-Former'}: Uses LLM tokens as queries with $P$ learnable key and value ``pooling tokens,'' each. Enabling variable-length outputs while acting as a bottleneck akin to dictionary learning.

\begin{figure*}[t]
\centering
    \includegraphics[width=\linewidth]{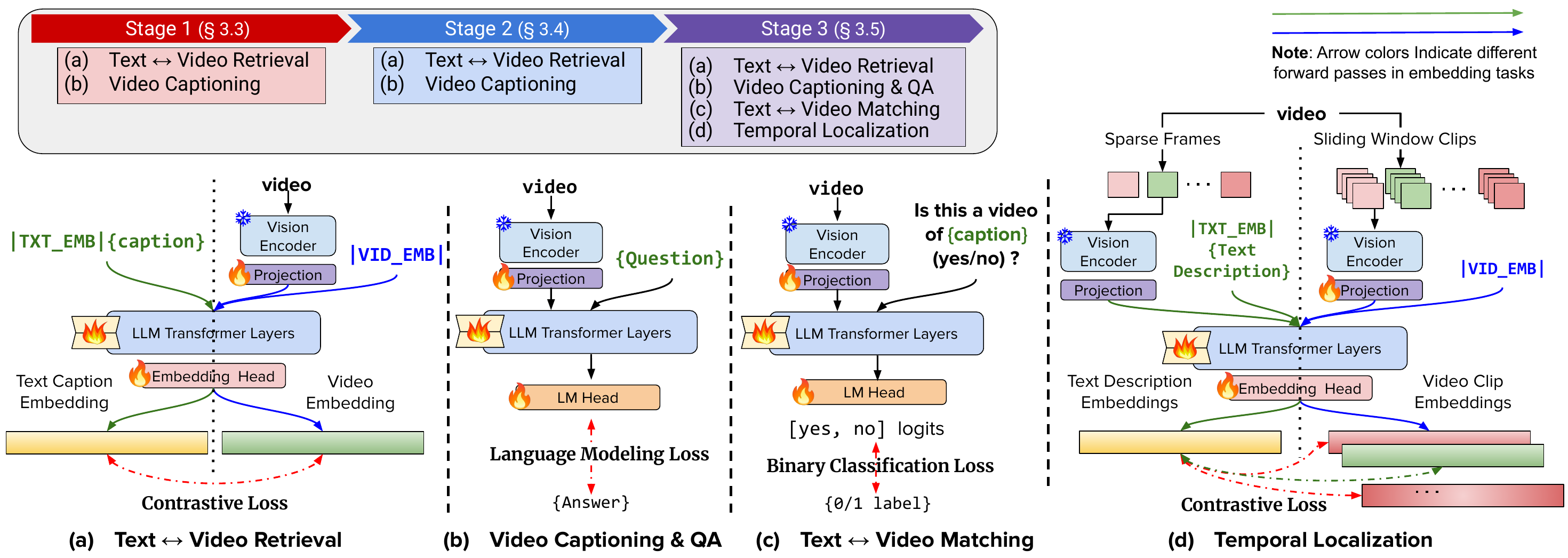}
    \vspace{-1.75em}
    \caption{Three stages of training. \textbf{Stage 1 (\autoref{sec:stage1})}, the large scale joint contrastive and generative pre-training, which utilizes (a) Video-Text retrieval task and (b) Video captioning on the large scale \texttt{Shutterstock} video  dataset. \textbf{Stage 2 (\autoref{sec:stage2})}, continues a similar joint pre-training approach (retrieval + captioning tasks as shown in (a) and (b)) on a smaller high quality dataset created using \texttt{Claude-3-Sonnet}. \textbf{Stage 3 (\autoref{sec:stage3})}, our multi-task fine-tuning stage, in addition to the prior tasks (retrieval + captioning) integrates (b) Video QA, (c) Video-Text matching and (d) Temporal Localization to improve all-around retrieval capabilities.}
    \label{fig:modes}
    \vspace{-1.5em}
\end{figure*}

Our embedding head (on the right in Fig.~\ref{fig:system}) receives tokens from the final LLM transformer layer and employs a learnable attentive pooling mechanism. Specifically, a transformer-based self-attention layer adaptively weighs tokens, using the LLM’s tokens as queries and $P$ learnable keys and values (``pooling tokens''). After this attention operation, the weighted representations are projected into the embedding space through a multi-layer perceptron (MLP) and subsequently mean-pooled. 

For the video embedding generation, our model uniquely leverages an end-of-sequence (\texttt{<EOS>})-triggered embedding generation mechanism. This mechanism allows the model to utilize variable inference steps tailored to each video's complexity - more steps for complex videos needing deeper analysis, fewer for simple ones. This adaptive computation approach substantially improves performance over existing fixed-step LLM-based embedding methods, offering both efficiency and improved representational quality. The proposed design is flexible and can be leveraged for various tasks, e.g. text and video embeddings for temporal localization, composed video retrieval etc. For text embedding generation, we rely on standard inference similar to prior works~\cite{grit}, with entire text embeddings being generated in a single step.

\vspace{-0.5em}
\subsection{Training Objectives}
\vspace{-0.5em}

We follow a three-stage training strategy: (1) pre-training on a large, diverse, and noisy dataset to jointly improve video captioning and embedding capabilities; (2) continued pre-training on a smaller, higher-quality dataset; and (3) multi-task finetuning to unlock the model’s performance across diverse downstream tasks. To ensure parameter efficiency during fine-tuning, we adopt Low-Rank Adapters (LoRA). The vision projection module and embedding head are fully trained to maximize representation learning. Our model is designed to effectively balance multimodal generative capabilities with state-of-the-art performance on embedding-oriented retrieval tasks. Across our training stages, a total of four different tasks are used, which are  described below.

\noindent\textbf{Text--Video Retrieval (Fig.~\ref{fig:modes}a).}
This task aligns video and text embeddings using CLIP-like~\cite{clip} contrastive loss, encouraging each paired example to outrank in-batch negatives.

\noindent\textbf{Video Captioning \& Question–Answering (Fig.~\ref{fig:modes}b).}
The goal of captioning is for the model to generate accurate descriptions of videos. The model learns to predict each token in the caption, given the previous tokens and the video. For QA, the model generates answers to questions about the video; the question serves as part of the prompt, guiding the model’s attention to answer accurately. We use the next-token prediction loss.

\noindent\textbf{Text--Video Matching (Fig.~\ref{fig:modes}c).}
As a complement to the retrieval task we also formulate a matching task where the model is trained to identify if a given video and caption form a matching pair. Binary decisions are supervised by labels.

\noindent\textbf{Temporal Localization (Fig.~\ref{fig:modes}d).}
Localization requires the model to match specific video segments to text annotations while distinguishing similar but incorrect segments. Sliding window hard negative mining provides challenging non-matching segments with limited temporal overlap (IoU $< 0.2$), helping the model learn to differentiate similar content. We use a standard contrastive formulation with one positive and two hard negatives per query.

\noindent\emph{Complete notation and full equations are provided in the Appendix~\ref{supp:notation}.}  The following subsections discuss our multi-stage training strategy in detail.

\vspace{-0.75em}
\subsection{Training Stage 1: Generative-Contrastive Pre-Training}
\label{sec:stage1}
\vspace{-0.5em}

In stage 1, we utilize a large video-caption dataset and jointly train the network using generative and contrastive supervision. In the first forward pass, for each video caption pair, video input tokens are passed along with a prompt asking the model to describe the video. Captioning output for the generative part and the video embedding are generated simultaneously in this pass. This is followed by a second computationally lighter forward pass with text-only input to generate the text embedding. The next token prediction loss is applied to the caption generation task, and the CLIP multi-modal contrastive loss is applied for the embedding generation task. The generative component aims to improve the model's ability to create coherent and contextually relevant textual descriptions of the video content, while the contrastive component enhances its discriminative power by teaching it to differentiate between relevant and irrelevant text-video pairs. This dual-objective training strategy ensures that the model not only learns to generate high-quality textual content but also develops a better alignment between modalities.

\vspace{-0.5em}
\subsection{Training Stage 2: Continued Pre-Training}
\vspace{-0.25em}
\label{sec:stage2}

While our pre-training data has a diverse collection of videos, the captions tend to be short summaries of the video content, often missing many details in the video. Since preserving important details from the video in our VideoLLM's feature space is beneficial for downstream tasks, we supplement our pre-training with a small intermediate pre-training phase using higher quality captions for a subset of videos from our pre-training set. The dataset for intermediate pre-training with high quality descriptive captions is discussed in detail in Section~\ref{sec:dataset}.


\vspace{-0.25em}
\subsection{Training Stage 3: Multi-Task Finetuning}
\vspace{-0.25em}

\label{sec:stage3}

In this stage, a high-quality finetuning dataset is used combining supervision across four tasks: captioning or question answering (QA), retrieval, matching and temporal localization. The fine-tuning process integrates these tasks to ensure that the model develops well-rounded capabilities, enabling it to handle diverse real-world scenarios that involve multi-modal understanding and generation. 

\vspace{-0.5em}
\section{Training Data}
\label{sec:dataset}
\vspace{-0.5em}

In this section, we provide details about the datasets used in the different training stages of our model. 
\vspace{-0.5em}
\subsection{Stage 1: Large Scale Pre-Training Data}
\vspace{-0.5em}

For pre-training, we use video–caption pairs from the licensed Shutterstock stock videos dataset\footnote{https://www.shutterstock.com/}, which contains 10M unique captions. Retaining one random clip per caption yields 10M video–caption pairs, equivalent to the WebVid-10M dataset~\cite{webvid} previously used in academic research.

\begin{figure}[h]
  \centering
  \includegraphics[width=\linewidth]{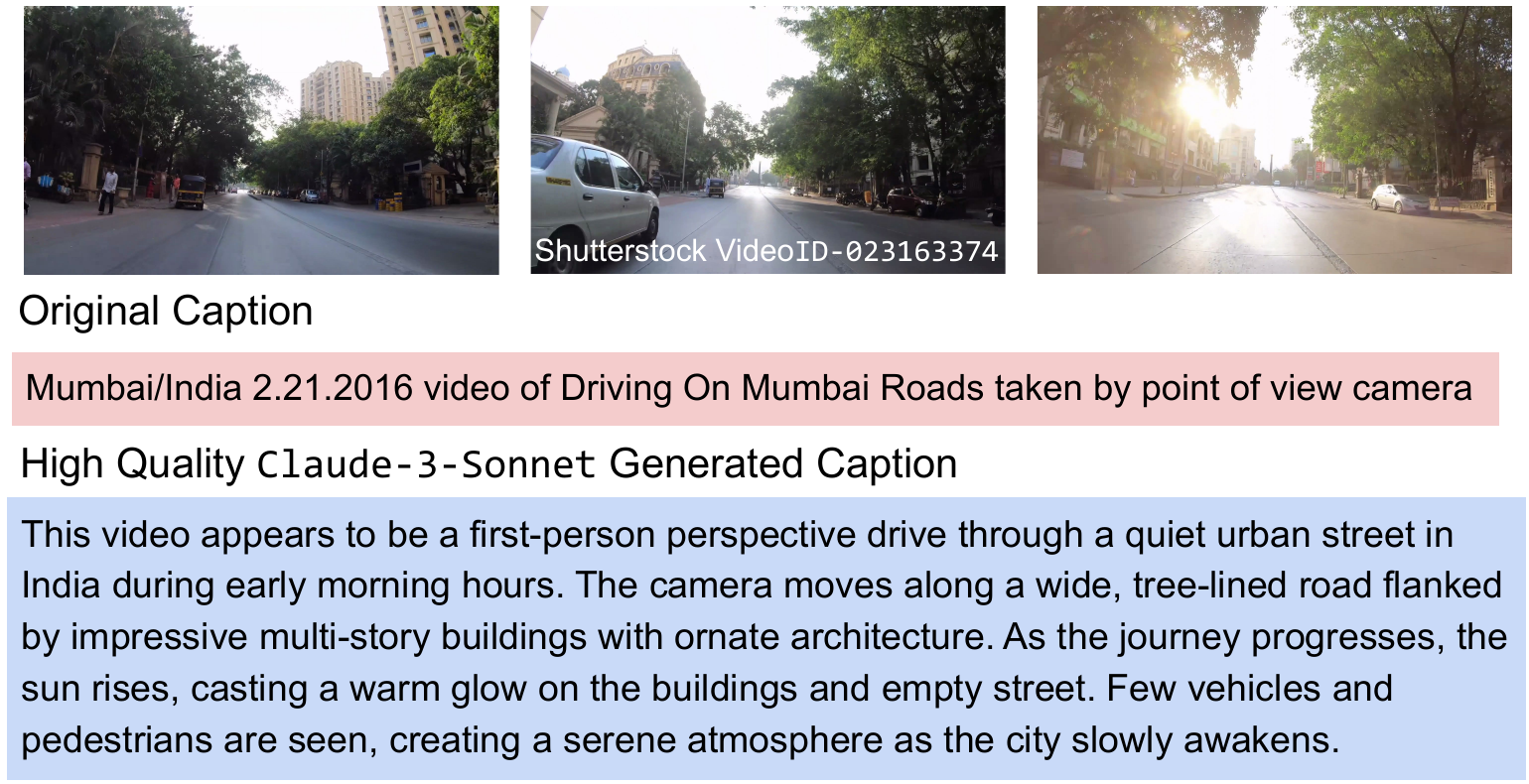}
  \vspace{-1.75em}
  \caption{Comparing original caption against \texttt{Claude-3-Sonnet} generated high-quality caption.}
  \label{fig:hqcaption}
  \vspace{-1.5em}
\end{figure}


\vspace{-0.5em}
\subsection{Stage 2: High Quality Intermediate Data}
\vspace{-0.5em}


  We use a re-captioned subset of the Shutterstock dataset, whose extensively labeled keywords we utilize for the balancing operation. We exclude keywords with $<$ 30 occurrences. Starting with the most frequent remaining keywords, we add a maximum of 500 videos per keyword to the candidate pool. Further details are in Supplementary~\ref{sec:datadist}. In order to generate this data within a limited budget, we generate captions for a diverse subset of 200,000 videos. We follow the insight from prior works (e.g., LLaVA~\cite{llava}, Meta-CLIP~\cite{metaclip}, Cambrian~\cite{cambrian}) that under-sampling common concepts is important for large scale vision-language training to achieve better generalization on rare concepts.

We use the \texttt{Claude-3-Sonnet} model prompted with 20 frames from the video, each resized to 480 pixels height, along with the following text prompt: 
\textit{``Provide detailed narrative caption for the video, covering its overall theme, visuals, characters, actions, scene, and key moments comprehensively"}. The original caption and generated high quality caption for a sample video are shown in Fig.~\ref{fig:hqcaption}.

\begin{table*}[t]
\centering
\setlength{\tabcolsep}{1pt}
\renewcommand{\arraystretch}{0.85}
\small
\begin{tabular}{
  p{4cm}
  >{\centering\arraybackslash}p{2cm}
  >{\centering\arraybackslash}p{2cm}
  >{\centering\arraybackslash}p{1.4cm}|
  >{\centering\arraybackslash}p{1.3cm}
  >{\centering\arraybackslash}p{1.3cm}
  >{\centering\arraybackslash}p{1.3cm}
}
\toprule
& \multicolumn{3}{c|}{\textbf{Text to Moment Retrieval}} & \multicolumn{3}{c}{\textbf{Text to Video Retrieval}}\\ 
\midrule
\textbf{Method} & \multicolumn{1}{c}{\textbf{ActivityNet}} & \multicolumn{1}{c}{\textbf{Charades}} & \multicolumn{1}{c|}{\textbf{QVHighlights}} & \multicolumn{1}{c}{\textbf{MSR-VTT}}  & \multicolumn{1}{c}{\textbf{DiDeMo}} & \multicolumn{1}{c}{\textbf{VATEX}} \\ 
& \multicolumn{1}{c}{\texttt{R@1,IoU=0.5}} & \multicolumn{1}{c}{\texttt{R@1,IoU=0.5}} & \multicolumn{1}{c|}{\texttt{mAP}} & \multicolumn{1}{c}{\texttt{R@1}} & \multicolumn{1}{c}{\texttt{R@1}} & \multicolumn{1}{c}{\texttt{R@1}}\\ 
\midrule
\multicolumn{4}{l|}{\textbf{Expert Models}} &   &   & \\
M-DETR~\texttt{\footnotesize{NeurIPS 21}} & - & 53.6\textsuperscript{\FT} & 35.7\textsuperscript{\FT} &  \xmark  &  \xmark  & \xmark \\ 
QD-DETR~\texttt{\footnotesize{CVPR 23}} & 33.2\textsuperscript{\FT} & 57.3\textsuperscript{\FT} & 38.9\textsuperscript{\FT} &   \xmark & \xmark   & \xmark \\ 
TR-DETR~\texttt{\footnotesize{AAAI 24}} & - & 57.6\textsuperscript{\FT} & 39.9\textsuperscript{\FT} &   \xmark & \xmark   & \xmark \\ 


SigLIP~\texttt{\footnotesize{ICCV 23}} & \xmark  & \xmark  & \xmark  & 51.7\textsuperscript{\FT} & 55.4\textsuperscript{\FT} & 40.8\\ 
VidLA~\texttt{\footnotesize{CVPR 24}} &  \xmark &  \xmark &  \xmark & 58.0\textsuperscript{\FT} & 61.1\textsuperscript{\FT} & -\\ 
InternVideo2-CLIP-1B~\texttt{\footnotesize{ECCV 24}} &  \xmark &  \xmark &  \xmark & 50.0 & 47.7 & 63.2\\ 

\midrule

\multicolumn{4}{l|}{\textbf{Video LLMs}} &   &   &   \\
Momentor-7B~\texttt{\footnotesize{ICML 24}} & 23.0 & 27.5 & 7.6 & \xmark  & \xmark  & \xmark \\ 
VTimeLLM-7B~\texttt{\footnotesize{CVPR 24}} & 27.8 & 27.5 & - & \xmark  & \xmark  & \xmark \\ 
TimeChat-7B~\texttt{\footnotesize{CVPR 24}} & 16.3 & 32.2 & 21.7 & \xmark   &\xmark   &  \xmark \\ 
Qwen2.5-VL-3B~\texttt{\footnotesize{arXiv}} & 28.6 & 38.1 & - & \xmark  &   \xmark & \xmark  \\ 
LLaVA-ST-7B~\texttt{\footnotesize{CVPR 25}} & 31.2 & 44.8 & - & \xmark   &\xmark   &  \xmark \\

\midrule
\textbf{ViLL-E-2.5B (Ours)} & \textbf{39.4}\scriptsize{\textcolor{britishracinggreen}{$\mathord{\uparrow}8.2$}} & \textbf{51.5}\scriptsize{\textcolor{britishracinggreen}{$\mathord{\uparrow}6.7$}} & 39.0 & \textbf{62.5}\scriptsize{\textcolor{britishracinggreen}{$\mathord{\uparrow}4.5$}} & \textbf{61.4} & \textbf{63.5}\\ 
\bottomrule
\end{tabular}
\vspace{-0.5em}
\caption{\textbf{Results on retrieval tasks}. In moment retrieval, we outperform specialized Video LLMs by over 8\%, and approach the performance of SotA dataset-specific expert models. In video retrieval we beat finetuned SotA models. \textsuperscript{\FT} models fine-tuned on train splits. \xmark - model is incapable of task}
\vspace{-2em}
\label{tab:sota}
\end{table*}

\vspace{-0.5em}
\subsection{Stage 3: Multi-Task Fine-Tuning Data}
\vspace{-0.5em}


Our fine-tuning stage is analogous to Instruction 
\setlength{\columnsep}{1em}
\setlength{\intextsep}{0pt}
\begin{wraptable}{r}{0.25\textwidth}
  \centering
  \setlength{\tabcolsep}{0.9pt}
  \renewcommand{\arraystretch}{0.95}
  
  \small
  \begin{tabular}{l|c|ccc}
    \toprule
    \textbf{Dataset} & \textbf{\#} & \textbf{Cap}  & \textbf{QA} & \textbf{Loc} \\
    \midrule
    MSR-VTT      & 10K & \Yes & \Yes & \xmark \\
    ActivityNet  & 30K & \Yes & \Yes & \Yes \\
    DiDeMo       & 10K & \Yes & \xmark  & \Yes \\
    Shutterstock & 50K & \Yes & \xmark  & \xmark \\
    \bottomrule
  \end{tabular}
  \vspace{-1em}
  \caption{Multi-Task Dataset.}
  \vspace{0.75em}
  \label{tab:iftmix}
\end{wraptable}

\vspace{-1em}
\noindent  FineTuning (IFT) in LLMs. We design our fine-tuning dataset to include multi-task supervision covering these tasks as well as video captioning. We select 100,000 samples from a small number of video-text datasets and obtain supervision from each as summarized in Table~\ref{tab:iftmix}. All these datasets contribute to captioning, retrieval and matching supervision. {\bf MSR-VTT} provides short video captions and question answering supervision. {\bf ActivityNet} and {\bf DiDeMo} consist of longer videos with distinct temporally localizable captions for chunks of the video. Hence, these datasets can provide localization supervision. {\bf ActivityNet} has also previously been annotated with question answering data. We add a random subset of Shutterstock video-caption pairs (original captions) to the mix to maintain long tail diversity.

\vspace{-0.5em}
\section{Results}
\label{sec:results}
\vspace{-0.5em}

First we demonstrate the unified capabilities of our model on 8 benchmark datasets. Next up, we demonstrate the unique capabilities of our model by solving novel tasks that are hard for any single  model prior to this work. Finally, we show ablations demonstrating the effectiveness of each component in our modeling and training approach.

\subsection{Retrieval Tasks}
\vspace{-0.5em}

We evaluate our model on video retrieval, and moment retrieval/temporal localization tasks. These benchmarks together provide a holistic evaluation of video understanding capabilities. We use standard evaluation protocols following prior works~\cite{timechat, vidla, videollava}.


\noindent \textbf{Moment Retrieval}: We test on ActivityNet-Captions, Charades-STA and QVHighlights. Contextualized text embeddings are extracted by passing {\small \texttt{\{sparse video frames\} <|txt\_embed|> \{text\}}}. Video clips are extracted using a  sliding window operation and clip embeddings are generated by passing {\small \texttt{\{clip frames\} <|vid\_embed|>}}. Text and Video Clip embedding similarities are post-processed using standard methods to get the matching segment.


\noindent \textbf{Video Retrieval}: We evaluate on MSR-VTT, DiDeMo and VATEX. Video embeddings are extracted by passing following input to the model: {\small\texttt{\{video\} <|vid\_embed|>}}, while text embeddings are extracted by passing {\small\texttt{<|txt\_embed|> \{text\}}}. Best pair is matched using nearest neighbor retrieval.



\noindent  In moment retrieval (Table~\ref{tab:sota}), ViLL-E outperforms localization specialized video-LLMs by nearly 10\%, and approaches or surpasses the per-dataset fine-tuned performance of non-LLM methods. In text-video retrieval, ViLL-E matches image (\texttt{SigLIP}) and video (\texttt{VidLA}) retrieval models.



\vspace{-0.5em}
\subsection{QA Tasks}
\vspace{-0.5em}

Among VideoLLMs specialized for temporal localization, our method achieves top performance in several categories (MSR, VCG, MVBench) indicated by green upward arrows showing improvements over previous methods. It slightly under-performs compared to LLaVA-ST on the MSVD dataset, indicated by a small red downward arrow. While SotA general VideoLLMs Qwen-2.5VL and LLaVA-Video outperform \texttt{ViLL-E} on MVBench \& VideoMME, our approach remains competitive.

\begin{table}[h]
  \centering
  \setlength{\tabcolsep}{0.5pt}
  \renewcommand{\arraystretch}{0.9}
  \small
  \begin{tabular}{lccccc}
    \toprule
     \textbf{Method}   & \textbf{MSR} & \textbf{VCG} & \textbf{MSVD} & \textbf{MVBench} & \textbf{VidMME}\\
    \midrule
    \multicolumn{5}{l}{\textbf{General Video LLMs}} \\
    Qwen-2.5VL & - & - & - & 67.0 & 61.5 \\
    LLaVA-Video & - & 3.5 & - & 58.6 & 63.3 \\
    \midrule
    \multicolumn{5}{l}{\textbf{Temporally-Specialized Video LLMs}} \\
    ST-LLM & 63.2 & 3.15 & 74.6 & 54.9 & 37.9 \\
    LLaVA-ST & 59.0 & 3.3 & \textbf{75.9} & 64.2 & \\
    Momentor & 55.6 & 3.0 & 68.9 & - & - \\
    VTimeLLM & 50.2 & 2.9 &  - &  - &  - \\
    TimeChat & 45.0 & \textit{2.3} & - & 38.5 & 34.7 \\
    \textbf{ViLL-E} & \textbf{65.2}
     & \textbf{3.7}
     & 75.2
     & \textbf{64.7} 
     & 45.0 \\
    \bottomrule
  \end{tabular}
  \vspace{-0.5em}
  \caption{ViLL-E is competitive with SotA VideoLLMs on QA tasks in both simple VideoQA (MSR‑VTT QA, VideoChatGPT Benchmark and MSVD‑QA) and Comprehensive Benchmarks (MVBench and VideoMME).}
  \label{tab:qa_sota}
  \vspace{-2em}
\end{table}

\vspace{-0.5em}
\subsection{Additional Tasks}
\vspace{-0.5em}
In this section we demonstrate ViLL-E's ability to solve some unique tasks that are made possible by having unified generative and embedding capabilities in one VideoLLM.  


\noindent \textbf{Two Stage Retrieval using Re-Ranking:} This task leverages the VideoLLM's reasoning capability to augment retrieval performance. In the first stage, Top-$K$ candidate videos matching a caption are retrieved using generated embeddings. During second stage matching, each candidate video-caption pair is input to the VideoLLM and the matching loss is used to re-rank the Top-$K$ candidate videos. The results in Table~\ref{tab:matching} ($K=25$) show that our two stage matching process results in an improvement of about 2\% in \texttt{R@1} accuracy.


\begin{table}[h]
  \centering
  \vspace{0.25em}
  \footnotesize
  \setlength{\tabcolsep}{3pt}
  \renewcommand{\arraystretch}{0.85}
  \begin{tabular}{l|cccc} 
    \toprule
    \textbf{MSR-VTT}  & \multicolumn{2}{c|}{T $\rightarrow$ V} & \multicolumn{2}{c}{V $\rightarrow$ T} \\ 
    \midrule
    \textbf{Method} & R@1 & R@5 & R@1 & R@5 \\ 
    \midrule
    VidLA          & 58.0          & \textbf{81.1} & 56.1          & 80.5 \\
    Ours (1 Stage) & 62.5          & 78.1          & 55.3          & 74.8 \\
    Ours (2 Stage) & \textbf{62.8} & 80.1          & \textbf{57.3} & \textbf{83.5} \\
    \bottomrule
  \end{tabular}
  \vspace{-1em}
  \caption{Two-stage retrieval + matching.}
  \label{tab:matching}
\end{table}

\noindent \textbf{Composed Video Retrieval}~\cite{covr} is the task of retrieving a video matching a given query video along with a modifier text. The retrieved video must closely match a hypothetical video containing the contents of the original video further modified by the instructions in the change text. This is a relatively new task in the field of video retrieval and forms a challenging zero-shot benchmark for our model. To solve this, we extract multi-modal query embeddings by passing {\small \texttt{\{source video frames\} <|txt\_embed|> \{change text\}}}, and retrieve videos by matching with target video embeddings extracted as we do for retrieval. The results in Table~\ref{tab:covr} demonstrate that in the zero-shot setting our method surpasses recent baselines by more than 5\%.


\begin{table}[h]
  \centering
  \footnotesize
  \vspace{0.25em}
  \setlength{\tabcolsep}{3pt}
  \renewcommand{\arraystretch}{0.9}
  \begin{tabular}{l|ccc} 
    \toprule
    \textbf{Method} & \textbf{R@1} & \textbf{R@5} & \textbf{R@10} \\ 
    \midrule
    COVR-BLIP~\texttt{\footnotesize{AAAI 24}}        & 45.46 & 70.46 & 79.54 \\ 
    Thawakar et al.~\texttt{\footnotesize{CVPR 24}}  & 47.52 & 72.18 & 82.37 \\ 
    Ours                                             & \textbf{53.13} & \textbf{74.80} & \textbf{85.95} \\
    \bottomrule
  \end{tabular}
  \vspace{-1em}
  \caption{Zero-Shot Composed Video Retrieval.}
  \vspace{-0.1em}
  \label{tab:covr}
\end{table}

\noindent \textbf{Detailed Video  Caption Retrieval:} Dual encoder video-text retrieval models, like CLIP, are limited by short context windows (e.g., 77 tokens). Our LLM approach supports much longer texts: \vspace{0.1em}

\begin{table}[h]
\centering
\footnotesize
\vspace{0.2em}
\setlength{\tabcolsep}{1.3pt}
\renewcommand{\arraystretch}{0.9}
\begin{tabular}{l|cccc|cccc} 
\toprule
\textbf{Method}    & \multicolumn{4}{c}{\textbf{Short Caption}} & \multicolumn{4}{|c}{\textbf{Long Caption}}                             \\ 
    & \multicolumn{4}{c}{\textbf{Avg.} = 27 words} & \multicolumn{4}{|c}{\textbf{Avg.} = 500 words }                             \\ 
\midrule
    & \multicolumn{2}{c|}{T $\rightarrow$ V}    & \multicolumn{2}{c|}{V $\rightarrow$ T} & \multicolumn{2}{c|}{T $\rightarrow$ V}    & \multicolumn{2}{c}{V $\rightarrow$ T}     \\ 
\midrule
    & \RetSymb{1}     & \RetSymb{5}   & \RetSymb{1}     & \RetSymb{5} & \RetSymb{1}     & \RetSymb{5}   & \RetSymb{1}     & \RetSymb{5}      \\ 
\midrule
SigLIP \texttt{ICCV23}   & 62.5 & 83.2 & 65.4 & 85.1 & 51.7 & 72.7 & 60.4 & 81.0 \\
VidLA \texttt{CVPR24}   & 61.2 & \textbf{85.7} & \textbf{65.9} & 85.0 & 45.3 & 65.2 & 55.9 & 72.7 \\
LongCLIP \texttt{ECCV24} & 63.0 & 85.4 & 61.5 & 83.6 & 73.5 & 91.7 & 74.5 & 92.8  \\ 
\textbf{ViLL-E}     & \textbf{64.3} & \textbf{85.7} & 65.5 & \textbf{85.4} & \textbf{75.7} & \textbf{92.4} & \textbf{75.1} & \textbf{93.3}     \\ 
\bottomrule
\end{tabular}
\vspace{-0.5em}
\caption{Zero-Shot Video Detailed Caption Retrieval.}
\label{tab:detailed}
\end{table}

\noindent thousands of tokens, enabling effective zero-shot retrieval using detailed captions. Existing benchmarks (e.g., YouCook2) are either too domain-specific or overlap with our training data. We use the AuroraCap-VideoDetailCaption dataset~\cite{auroravdc}, which includes both long and short captions for 1,027 videos. As shown in Table~\ref{tab:detailed}, our model benefits from longer captions, improving \texttt{R@1} by 11.4\% (text-to-video) and 9.6\% (video-to-text). In contrast, models like SigLIP~\cite{siglip} and VidLA~\cite{vidla} are constrained by short text encoder contexts (64 tokens). LongCLIP~\cite{longclip}, with a 248-token limit, improves on long captions but lags our approach.

\begin{table}[h]
\centering

\begin{subtable}[h]{0.5\textwidth}
\centering
\footnotesize
\setlength{\tabcolsep}{0.5pt}
\renewcommand{\arraystretch}{0.9}
\begin{tabular}{@{}ccc|cc|ccc|ccc@{}}
\toprule
   G & C & M & MSR & VCG & MSR & DDM & VTX & ANet & QVH & Ch. \\ 
\midrule
   \checkmark & \checkmark & \checkmark & 65.1 & 3.7 & 62.8 & 61.5 & 63.7 & 39.4 & 38.9 & 51.7 \\
 \midrule
   \checkmark & \checkmark & \texttimes & 63.9 & 3.7 & 60.3 & 60.2 & 62.2 & 39.1 & 38.5 & 51.5 \\
   \checkmark & \texttimes & \texttimes & 61.3 & 3.0 & 25.1 & 23.5 & 11.9 & 28.7 & 23.8 & 42.1 \\
   \texttimes & \checkmark & \texttimes & 45.5 & 2.1 & 54.7 & 53.1 & 55.2 & 29.3 & 30.3 & 48.8 \\
\bottomrule
\end{tabular}
\vspace{-0.5em}
\caption{Supervision Type (During Finetuning)}
\vspace{0.3em}
\end{subtable}
\begin{subtable}[h]{0.49\textwidth}
\centering
\footnotesize
\setlength{\tabcolsep}{0.5pt}
\renewcommand{\arraystretch}{0.9}
\begin{tabular}{@{}c|cc|ccc|ccc@{}}
\toprule
 Head & MSR & VCG & MSR & DDM & VTX & ANet & QVH & Ch. \\

  \midrule
    Ours  & 55.9 & 3.1 & 49.3 & 45.5 & 45.3 & 32.3 & 32.6 & 49.3 \\
    \midrule
    \texttimes & 40.8 & 2.8 & 30.7 & 29.1  & 32.1 & 19.8 & 20.4 & 30.7   \\
    Linear & 42.9 & 2.8 & 32.4 & 29.3 & 33.2 & 23.6 & 25.4 & 42.7    \\
    MLP  & 51.7 & 3.0 & 43.5 & 39.8 & 39.5 & 30.2 & 29.7 & 47.1\\
    Self-Att & 52.5 & 3.2 & 43.8 &  40.9 & 40.9 & & 28.2 & 45.7\\
    Q-Former & 52.1 & 3.0 & 47.5 & 43.2 & 42.8 & & 27.9 & 44.2 \\
    K-Former & 52.7 & 3.1 & 49.0 & 44.1 & 44.7 & & 32.8 & 47.4 \\
\bottomrule
\end{tabular}
\vspace{-0.5em}
\caption{Embedding Head Design}
\vspace{0.3em}
\end{subtable}
\begin{subtable}[h]{0.48\textwidth}
\centering
\footnotesize
\setlength{\tabcolsep}{0.5pt}
\renewcommand{\arraystretch}{0.9}
\begin{tabular}{@{}cc|cc|ccc|ccc@{}}
\toprule
  Pre & HQ  & MSR & VCG & MSR & DDM & VTX & ANet & QVH & Ch. \\
\midrule
  \checkmark & \checkmark  & 65.1 & 3.7 & 62.8 & 61.5 & 63.7 & 39.4 & 38.9 & 51.7 \\
 \midrule
  \checkmark & \texttimes  & 63.2 & 3.3 & 58.6 & 59.6 & 60.5 & 36.8 & 38.8 & 51.9 \\
  \texttimes & \texttimes  & 55.9 & 3.1 & 49.3 & 45.5 & 45.3 & 32.3 & 32.6 & 49.3 \\
\bottomrule
\end{tabular}
\vspace{-0.5em}
\caption{Impact of Pre-Training}
\vspace{0.5em}
\end{subtable}
\vspace{-0.5em}
\vspace{-0.5em}
\caption{\textbf{Ablation Experiments.}}
\vspace{0.25em}
\begin{itemize}[nosep,leftmargin=0pt,label={}]\footnotesize
  \item \textbf{Benchmarks (left to right).} \textbf{QA}: MSR‑VTT‑QA, VideoChatGPT. \textbf{Retrieval}: MSR‑VTT, DiDeMo, VATEX. \textbf{Localization}: ActivityNet, QVHighlights, Charades‑STA.
  \item \textbf{Legend:} \textbf{(a)} G = Generative, C = Contrastive, M = Matching. \textbf{(c)} \textit{Pre.} = 10M scale; \textit{HQ} = High‑Quality 200k.
  \end{itemize}
\label{tab:ablations}
\vspace{-1.5em}

\end{table}

\noindent


\vspace{-0.5em}
\subsection{Ablations}
\label{sec:ablations}
\vspace{-0.5em}
We ablate the design choices for key components of our approach. For computational efficiency, some experiments are performed without pre-training.



\noindent \textbf{Supervision Types:} (Table~\ref{tab:ablations}a) We find that combining generative and contrastive objectives during fine-tuning complements each other, resulting in improvement on both generative and embeddings tasks. Retrieval and localization performances drop moderately on removing matching loss, and significantly on removing contrastive training, indicating the importance of learning good embeddings. 


\noindent \textbf{Embedding Head} (Table~\ref{tab:ablations}b) We find that our attentive pooling head outperforms other potential approaches discussed previously in Section~\ref{sec:embedhead}.

\noindent \textbf{Pre-Training} (Table~\ref{tab:ablations}c) The large scale pre-training stage is very important for retrieval tasks, whereas the second pre-training stage with longer higher quality captions helps particularly in longer video datasets and generative tasks.






\noindent Additional experiments and analysis of our adaptive embedding mechanism: including inference latency, per-duration-bin retrieval accuracy, and distribution of generated token counts, are provided in Appendix~\ref{sec:extrabl} (Tables~\ref{tab:latency} and \ref{tab:duration}).

\vspace{-0.5em}
\section{Conclusion}
\vspace{-0.5em}

We introduced ViLL-E, a unified VideoLLM that combines text generation and embedding-based retrieval within one model. Through a multi-stage generative–contrastive training strategy and an adaptive embedding head, ViLL-E achieves strong results across retrieval, localization, and QA, while enabling new zero-shot tasks like composed and long-caption retrieval. This work highlights the potential of merging generative and discriminative learning for unified multimodal understanding.

\section{Limitations}

Our model inherits some limitations of the base Paligemma LLM, e.g. it lacks general multi-turn conversation abilities. Secondly, because our training dataset is primarily in English, we expect that our model would lose some of Paligemma's multi-lingual capabilities. Our work serves as a proof of concept for the unification of generative and embedding capabilties in VideoLLMs, and any practical model would need to address the issues of multi-turn conversations and multi-linguality.


{
    \small
    \bibliography{custom}
}

\clearpage
\appendix


\section*{Overview of Supplementary Material}

\begin{enumerate}[label=\textbf{\Alph*.}, itemsep=0.75em, parsep=0pt, leftmargin=*, align=left]

\item Additional Ablations

\item Loss Notation and Equations

\item Implementation Details

\item Broader Impacts

\item Privacy Safeguards

\item Licensing Information

\item Additional Details About Localization Inference

\item Additional Details About Two-stage retrieval Inference

\item Additional Details About Composed Video Retrieval Inference

\item Additional Details About Stage 1 joint pre-training

\item Additional Details About Stage 3 Temporal Localization training

\item Details About High Quality Pretraining Dataset Construction: Balancing

\item Visualizing High Quality Dataset Statistics

\item Qualitative Captioning Results

\item Visualization of Video Embeddings

\end{enumerate}










\section{Additional Ablations}
\label{sec:extrabl}

\noindent \textbf{Number of Pooling Tokens} (Table~\ref{tab:pooling}) We find that in practice, a relatively high value of (we use 256) is optimal (compare to Q-Former, where usually 32 queries are used). Going beyond 256 to higher values, gains seem minimal.

\begin{table}[h]
\centering
\footnotesize
\setlength{\tabcolsep}{2pt}
\renewcommand{\arraystretch}{0.9}
\begin{tabular}{@{}cc|c|cc|c@{}}
\toprule
 \# & Pre. & VCG & MSR & VTX  & QVH  \\
  \midrule
    Ours (256) & \texttimes  & 3.1 & 49.3 & 45.3 & 32.6 \\
    \midrule
    32   & \texttimes &  2.7	& 34.8	& 35.5	& 25.7 \\
    64  & \texttimes  & 	3.0	& 42.5	& 39.2	& 29.5 \\
     128  & \texttimes  & 	3.1	& 48.6	& 44.9	& 32.0 \\
      512  & \texttimes  & 3.1	& 49.5	& 45.0	& 32.4
\\
\bottomrule
\end{tabular}
\vspace{-0.5em}
\caption{Number of Pooling Tokens}
\label{tab:pooling}
\vspace{0.5em}
\end{table}

\noindent \textbf{Number of Embedding Tokens} (Table~\ref{tab:tokens}) Our video embedding generation outputs a variable number of tokens (until \texttt{<EOS>}) to send to the embedding head. Whereas prior encoder approaches (e.g. GRIT~\cite{grit}) generate a fixed number of tokens, to match this setting and measure the benefit of our dynamic approach we also train the model to use just 1 (or 5) token for embedding generation.

\begin{table}[h]
\vspace{0.5em}
\centering
\footnotesize
\setlength{\tabcolsep}{0.5pt}
\renewcommand{\arraystretch}{0.9}
\begin{tabular}{@{}cc|cc|ccc|ccc@{}}
\toprule
 \# & Pre. & MSR & VCG & MSR & DDM & VTX & ANet & QVH & Ch. \\
  \midrule
    Ours & \texttimes  & 55.9 & 3.1 & 49.3 & 45.5 & 45.3 & 32.3 & 32.6 & 49.3 \\
    \midrule
    1   & \texttimes & 50.2 & 2.6 & 45.8 & 37.6 & 38.9 & 23.7 & 24.1 & 36.8 \\
    5  & \texttimes  & 54.1 & 2.8 & 47.9 & 40.3 & 41.2 & 27.3 & 27.8 & 42.3\\
\bottomrule
\end{tabular}
\vspace{-0.5em}
\caption{Number of Embedding Tokens}
\label{tab:tokens}
\end{table}

\noindent \textbf{Inference Latency vs.\ Number of Embedding Tokens} (Table~\ref{tab:latency}) We measure inference latency on an AWS \texttt{g6e.8xlarge} instance equipped with an NVIDIA L40S GPU. Our adaptive method adds only marginal overhead compared to a fixed 5-token baseline, while achieving substantially better accuracy (Table~\ref{tab:tokens}).

\begin{table}[h]
\vspace{0.5em}
\centering
\footnotesize
\setlength{\tabcolsep}{6pt}
\renewcommand{\arraystretch}{0.9}
\begin{tabular}{@{}lc@{}}
\toprule
\textbf{Tokens} & \textbf{Latency (ms)} \\
\midrule
Fixed -- 1  & 238 \\
Fixed -- 5  & 257 \\
Fixed -- 10 & 326 \\
Adaptive (Ours) & 262 \\
\bottomrule
\end{tabular}
\vspace{-0.5em}
\caption{Inference latency vs.\ number of embedding tokens for 12-frames (batch size 16) on NVIDIA L40S.}
\label{tab:latency}
\end{table}

\noindent \textbf{Adaptive Tokens Across Video Durations} (Table~\ref{tab:duration}) To further analyze the adaptive token mechanism, we break down the ActivityNet Captions dataset by video duration. This dataset has a wide and nearly uniform distribution of videos up to 5 minutes, providing balanced bins. Our adaptive method generalizes better to variable video lengths, significantly outperforming fixed-token approaches across all duration bins.

\begin{table}[h]
\vspace{0.5em}
\centering
\footnotesize
\setlength{\tabcolsep}{3pt}
\renewcommand{\arraystretch}{0.9}
\begin{tabular}{@{}lccccc@{}}
\toprule
\textbf{Method} & \textbf{0--60s} & \textbf{60--120s} & \textbf{120--180s} & \textbf{180s+} & \textbf{Overall} \\
\midrule
VidLA Baseline     & 65.7 & 66.5 & 65.0 & 63.5 & 65.2 \\
Fixed Token = 1    & 66.2 & 64.7 & 63.5 & 62.2 & 64.1 \\
Fixed Tokens = 5   & 66.6 & 67.4 & 66.4 & 65.4 & 66.5 \\
Adaptive (Ours)    & \textbf{67.9} & \textbf{68.5} & \textbf{67.8} & \textbf{67.1} & \textbf{67.9} \\
\bottomrule
\end{tabular}
\vspace{-0.5em}
\caption{ActivityNet Captions Text-to-Video Retrieval (R@1) broken down by video duration.}
\label{tab:duration}
\end{table}

\noindent To verify that the model actually leverages variable-length generation, we measure the average number of tokens produced by the adaptive method per duration bin: \textbf{0--60s}: 8.6 tokens; \textbf{60--120s}: 10.1 tokens; \textbf{120--180s}: 11.5 tokens; \textbf{180s+}: 14.8 tokens. This confirms that the model naturally ``thinks longer'' for more complex, longer videos.

\section{Shared Notation}
\label{supp:notation}

For clarity we reserve a single symbol for each recurring concept.  
An \emph{entire video} is denoted by a capital letter, \(V\), whereas a \emph{temporal segment} cut from that video is represented in lowercase, \(v\). Hard‑negative segments for localization are denoted by \(v_{ij}^{n}\), where \(j\) indexes the negatives for sample \(i\). Any natural‑language sequence—be it a caption, a question, or an answer—is
denoted by \(t\); its \(k\)-th token is \(w_k\), and the total length is \(L_{\text{cap}}\) for captions or \(L_{\text{ans}}\) for answers.
The LLM maps raw inputs to embeddings, producing
\(\mathbf{e}_V = f(V,<m>)\), \(\mathbf{e}_v = f(v,<m>)\) and
\(\mathbf{e}_t = f(t,<m>)\), $f(.)$ representing our model and $<m>$ represents the mode token, e.g. \texttt{|VID\_EMB|} is the video embedding token. All contrastive objectives use the same similarity function
\(\operatorname{sim}(\cdot,\cdot)\) (cosine by default) scaled by a temperature
parameter \(\tau\). We use a mini‑batch of size \(N\).

\vspace{0.6em}
\noindent\textbf{Text--Video Retrieval (Fig.~\ref{fig:modes}a).} This task aligns video and text embeddings using CLIP-like~\cite{clip} contrastive loss.
\vspace{-0.75em}
\[
\mathcal{L}_{\text{ret}}
= -\frac{1}{N}\sum_{i=1}^{N}
  \log\frac{\exp\bigl(\operatorname{sim}(\mathbf{e}_V^{\,i},\mathbf{e}_t^{\,i})/\tau\bigr)}
           {\sum_{j=1}^{N}\exp\bigl(\operatorname{sim}(\mathbf{e}_V^{\,i},\mathbf{e}_t^{\,j})/\tau\bigr)} .
\]
\vspace{-1em}

\noindent\textbf{Video Captioning \& Question–Answering (Fig.~\ref{fig:modes}b).} The goal of captioning is for the model to generate accurate descriptions of videos. The model learns to predict each token in the caption, given the previous tokens and the video. Whereas for QA, the model generates answers to questions about the video. The question serves as part of the prompt, guiding the model’s attention to answer accurately.The next-token prediction loss is used.
\vspace{-1em}
\[
\begin{aligned}
\mathcal{L}_{\text{cap}} &= -\sum_{k=1}^{L_{\text{cap}}}
      \log p\bigl(w_k^{\text{cap}}\mid w_{<k}^{\text{cap}},\,V\bigr),
\end{aligned}
\]\vspace{-1em}
\[
\begin{aligned}
\mathcal{L}_{\text{qa}} &= -\sum_{k=1}^{L_{\text{ans}}}
      \log p\bigl(w_k^{\text{ans}}\mid w_{<k}^{\text{ans}},\,V,\,t_{\text{ques}}\bigr).
\end{aligned}
\]

\noindent\textbf{Text--Video Matching (Fig.~\ref{fig:modes}c).} As a complement to the retrieval task we also formulate a matching task where the model is trained to identify if a given video and caption form a matching pair. Binary decisions are supervised by the label \(y\in\{0,1\}\).
\vspace{-0.5em}
\[
\mathcal{L}_{\text{m}}
= -\bigl(
  y\log p(\text{Y}\mid V,t)
  + (1-y)\log p(\text{N}\mid V,t)
\bigr).
\]
\vspace{-1.5em}

\noindent\textbf{Temporal Localization (Fig.~\ref{fig:modes}d).}
Localization requires the model to match specific video segments to text annotations while distinguishing similar but incorrect segments. Sliding window hard negative mining provides challenging non-matching segments with limited temporal overlap (IoU $<$ 0.2), helping the model learn to differentiate closely related content.  Let $s_i^{+} \triangleq \operatorname{sim}(\mathbf e_v^{\,i}, \mathbf e_t^{\,i})$ (positive) and $s_{ij}^{-} \triangleq \operatorname{sim}(\mathbf e_v^{\,i,j}, \mathbf e_t^{\,i})$ (j-th hard negative for sample i, $j\in\mathcal N_i$; IoU $<0.2$). The localization loss, \( \mathcal{L}_{\text{l}} \), can be written as
\begin{equation}
\mathcal L_{\text{l}}
= -\frac{1}{N}\sum_{i=1}^{N}
\log \frac{\exp\!\bigl(s_i^{+}/\tau\bigr)}
{\exp\!\bigl(s_i^{+}/\tau\bigr) + \sum_{j\in\mathcal N_i}\exp\!\bigl(s_{ij}^{-}/\tau\bigr)} .
\end{equation}

\vspace{-1em}

\section{Implementation Details and Hyperparameters}
\label{sec:implement}

We use PaliGemma as our primary MLLM backbone. PaliGemma is openly licensed (Apache 2.0) and is based on the Gemma-2B language model and uses SigLIP-SO-400M ViT as its vision encoder. The single layer embedding head has the same dimensions as a Gemma-2B layer (\texttt{hidden\_dim} = 2048, \texttt{mlp\_dim} = 16384, and \texttt{n\_heads} = 8). All video frames are resized to 224 $\times$ 224 before feeding into the model. LoRA adapters of rank 128 were used for finetuning and are applied to all linear layers of the LLM transformer (i.e., excluding the vision encoder, embedding layers, and output heads). PaliGemma's \texttt{<unused0>} and \texttt{<unused1>} tokens are leveraged to initialize \texttt{<txt\_embed>} and \texttt{<video\_embed>}. A modified Huggingface \texttt{Trainer} is used for training, with \texttt{FSDP} mode used across GPUs to reduce $VRAM$ usage. Training hyperparameters for each stage are provided in Table~\ref{tab:hyperparameters}. The key reason behind these choices is improving the efficiency of the large scale Stage 1 training. The learning rates are chosen on the basis of small training runs on the MSR-VTT dataset. Inference latency is benchmarked on an AWS \texttt{g6e.8xlarge} instance with an NVIDIA L40S GPU. For temporal localization, we use a sliding window of 10 seconds with a stride of 5 seconds (see Appendix~G).

\noindent \textbf{Multi-Task Finetuning Batch Creation}: Following Table~\ref{tab:iftmix}, each video in our dataset has 4 boolean flags indicating which training tasks are supported for the video. When a random batch of videos is sampled, the flags are sampled with them. The flags are used to filter a partial task specific minibatch, with one minibatch per task. Gradients are accumulated across all training tasks before updating the model weights.

\begin{table}[htbp]
\centering
\setlength{\tabcolsep}{3pt}
\renewcommand{\arraystretch}{0.85}
 \setlength{\tabcolsep}{2pt}
\begin{tabular}{lcc}
\toprule
\textbf{Hyperparameter} & \textbf{Stage 1/2} & \textbf{Stage 3} \\
\midrule
Number of GPUs & 32 & 8 \\
Batch Size (per GPU) & 12 & 8 \\
Gradient Accumulation & 2 & 4 \\
Optimizer & Lion & AdamW \\
Base Learning Rate & 1e-5 & 5e-6 \\
Warmup Steps & 2000 & 1000 \\
Weight Decay$^\ast$ & 0.00001 & - \\
LoRA Rank & 128 & 32 \\
Training Steps & 25,000 & 4,000 \\
Max Token Length & 2304/4096  & 4096 \\
\bottomrule
\end{tabular}
\caption{Training hyperparameters across different stages. $^\ast$- only applied to new embedding head.}
\label{tab:hyperparameters}
\end{table}

\section{Broader Impacts}

A key finding of our work is the ability to unify generative and embedding models, which could potentially reduce the amount of energy used and environmental impact from maintaining two different model pipelines for inference in practical applications. In terms of potential negative impact, our model is very general, and hence shares some of the potential for misuse inherent to generalist computer vision models.

\section{Privacy Safeguards}

 All videos are stored and used with the faces of humans blurred using the \texttt{deface} python package. 

\section{Licenses}


\textbf{WebVid-10M (Shutterstock clips).}
A proprietary corpus governed by Shutterstock’s Terms of Service; URLs and captions may be downloaded for \emph{internal, non-commercial research} only, and redistribution is forbidden. 

\textbf{MSR-VTT.}
Released by Microsoft Research solely for academic research;
no explicit open-source license is attached, so usage is restricted to the terms on the dataset website/publication. 

\textbf{ActivityNet.}
Distributed under the MIT License, permitting commercial and non-commercial use provided that copyright and license notices are retained. 

\textbf{DiDeMo.}
Available under the BSD 2-Clause license, a permissive license that allows modification and redistribution with minimal obligations. 

\textbf{VATEX.}
Shared under Creative Commons Attribution 4.0 International (CC BY 4.0); free use with attribution. 

\textbf{QVHighlights.}
Released with a Creative Commons Attribution–NonCommercial–ShareAlike 4.0 license (CC BY-NC-SA 4.0); commercial use is prohibited and derivatives must adopt the same license. 

\textbf{Charades \& Charades-STA.}
Allen AI provides these datasets under a \emph{non-commercial research} license; any for-profit use requires separate permission. 


\textbf{SigLIP-SO-400M visual encoder.}
Model weights and code are released under the Apache License 2.0. 

\textbf{PaliGemma-3B multimodal LLM.}
Weights are covered by Google’s \emph{Gemma Open Model License 1.0}; accompanying reference code is Apache 2.0. 

\textbf{Claude-3 Sonnet (caption generator).}
A proprietary model whose use is governed by Anthropic’s Terms of Service. 





\section{Localization Inference}

We apply post-processing inspired from prior embedding-based localization approaches to merge segments.

We split the video into non-overlapping windows and compute a text-video similarity score for each segment. We take the highest-scoring window (with score $s_{\text{seed}}$) as the seed, then merge adjacent windows to the left and right as long as each neighbor’s score exceeds a merge threshold ($\tau_{\text{merge}}$) or is at least a fixed ratio (i.e., $\alpha$) of the seed’s score. Formally, a neighboring window $i$ is merged if
\[
s_i \ge \tau_{\text{merge}} \quad \text{or} \quad s_i \ge \alpha \cdot s_{\text{seed}}.
\]
Expansion stops at the first neighbor on either side that fails both conditions. The final prediction is the union of all accepted windows. Center correction is applied to the start and end clips if they are not the seed clip.

Some prior works that we referred to in order to develop our post-processing strategy include
\cite{zhang2022actionformer, liu2022endtoend}.

We ablate some of the choices related to our localization inference protocol here:

\begin{table}[h]
\centering
\begin{tabular}{cccc}
\toprule
\textbf{Window} & \textbf{Stride} & \textbf{IoU} & \textbf{R@1, IoU=0.5} \\
\midrule
5s & 5s & 0.3 & 39.0 \\
5s & 5s & 0.4 & 39.4 \\
5s & 5s & 0.5 & 35.7 \\
10s & 10s & 0.4 & 39.2 \\
\textbf{10s} & \textbf{5s} & \textbf{0.4} & \textbf{41.7} \\
15s & 10s & 0.4 & 39.9 \\
\bottomrule
\end{tabular}
\caption{Ablation of localization inference protocol choices. Score on ActivityNet Dataset.}
\end{table}

\section{Matching Based Two Stage Retrieval}
\label{sec:matching}

We provide more details of how we enhance video-caption retrieval performance by combining retrieval techniques with advanced matching capabilities of a VideoLLM. Below, we describe the steps for each stage:

\noindent \textbf{Stage 1: Embedding-Based Retrieval}

\begin{enumerate}
    \item \textbf{Embedding Generation:}
    \begin{itemize}
        \item Text embeddings for the input caption are generated using ViLL-E in text embedding mode.
        \item Video embeddings for all videos in the database are pre-computed using ViLL-E in video embedding mode.
    \end{itemize}

    \item \textbf{Similarity Computation:}
    \begin{itemize}
        \item The similarity between the caption embedding and each video embedding is computed, typically using cosine similarity.
    \end{itemize}

    \item \textbf{Top-$K$ Retrieval:}
    \begin{itemize}
        \item The $K$ most similar video candidates are selected based on the similarity scores. This reduces the search space for the more computationally intensive second stage.
    \end{itemize}
\end{enumerate}

\begin{figure*}[t]
    \centering
    \includegraphics[width=\linewidth]{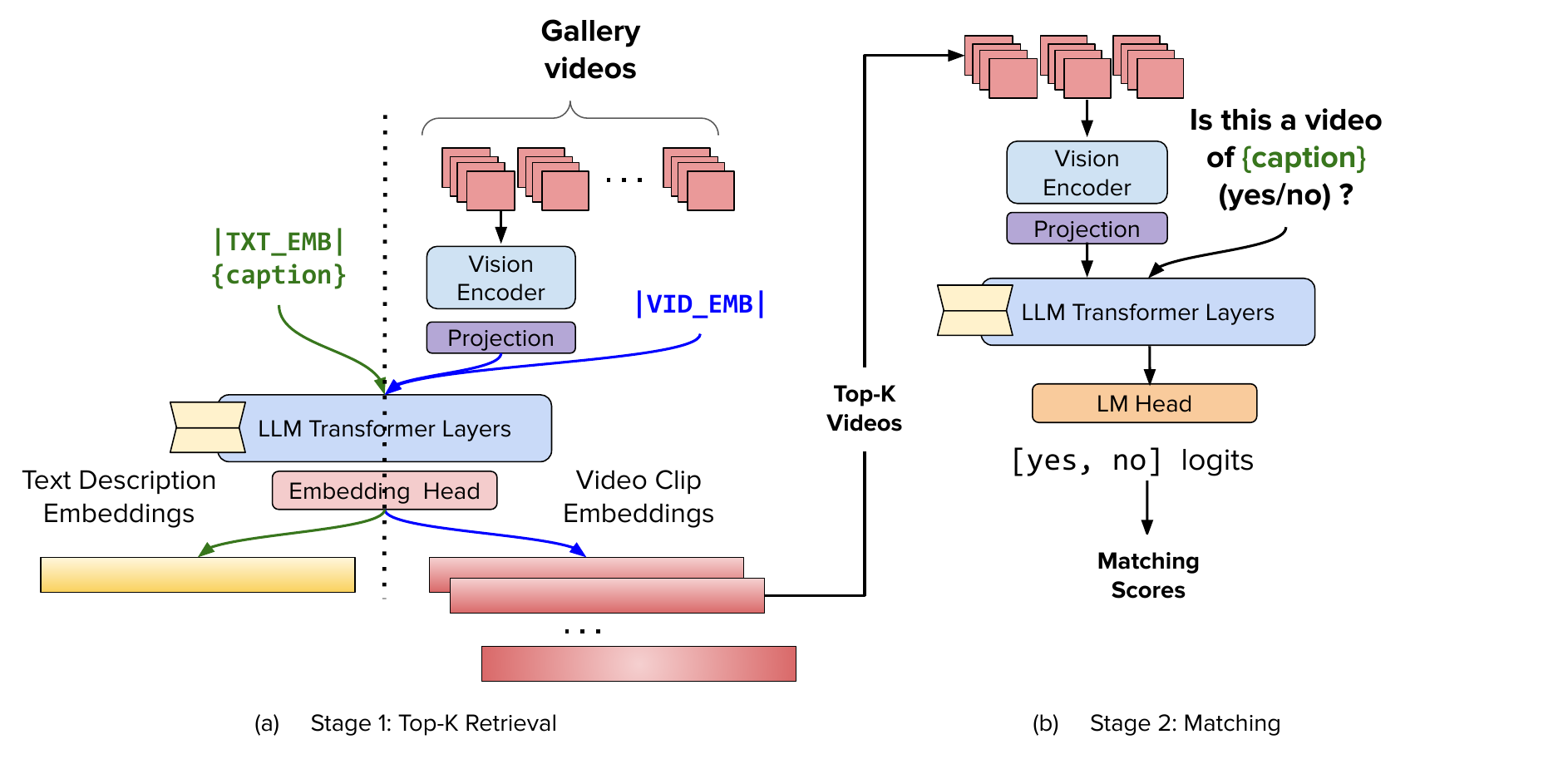}
    \caption{Two Step Retrieval Inference.}
    \label{fig:twostep}
    \vspace{-1em}
\end{figure*}

\noindent \textbf{Stage 2: Matching-Based Re-Ranking}

\begin{enumerate}
    \item \textbf{Video-Caption Pair Processing:}
    \begin{itemize}
        \item Each video in the Top-$K$ set is paired with the input caption and fed into the VideoLLM.
        \item A carefully designed prompt is used to guide the VideoLLM in determining whether the pair is a match.
    \end{itemize}

    \item \textbf{Prediction and Scoring:}
    \begin{itemize}
        \item For each video-caption pair, the VideoLLM predicts a likelihood score for both \texttt{Yes} (match) and \texttt{No} (no match) tokens.
        \item A matching score is computed from the logits of these predictions (e.g., softmax probabilities or logit differences).
    \end{itemize}

    \item \textbf{Re-ranking:}
    \begin{itemize}
        \item The Top-$K$ ($K=20$) videos are re-ranked based on their matching scores, with higher scores indicating stronger matches.
    \end{itemize}
\end{enumerate}

\noindent \textbf{Benefits of the Two-Stage Approach}

\begin{itemize}
    \item \textbf{Efficiency:} By leveraging computationally efficient embedding-based retrieval in the first stage, the system avoids evaluating all database videos
    \item \textbf{Accuracy:} The VideoLLM’s reasoning capabilities in the second stage allow it to resolve ambiguities and refine rankings, leading to a significant improvement in retrieval accuracy (e.g., $\sim2\%$ increase in \texttt{R@1} accuracy).
    \item \textbf{Scalability:} The division into two stages ensures that the system can scale to large databases without compromising on retrieval quality.
\end{itemize}

\noindent \textbf{Prompt for Matching}

The prompt provided to the VideoLLM for second-stage matching is structured as follows:

\begin{itemize}
    \item \textbf{Input:}
    \texttt{\textless Video Tokens\textgreater} Caption: \texttt{[Retrieved first stage caption]}. Question: Does the above video match the caption? Answer "Yes" or "No".

    \item \textbf{Output:}
    \begin{itemize}
        \item Token logits for \texttt{Yes} and \texttt{No}, which are further used to compute the matching score.
    \end{itemize}
\end{itemize}

\section{Composed Video Retrieval}
\label{sec:covr}

\begin{figure*}[t]
    \centering
    \includegraphics[width=0.65\linewidth]{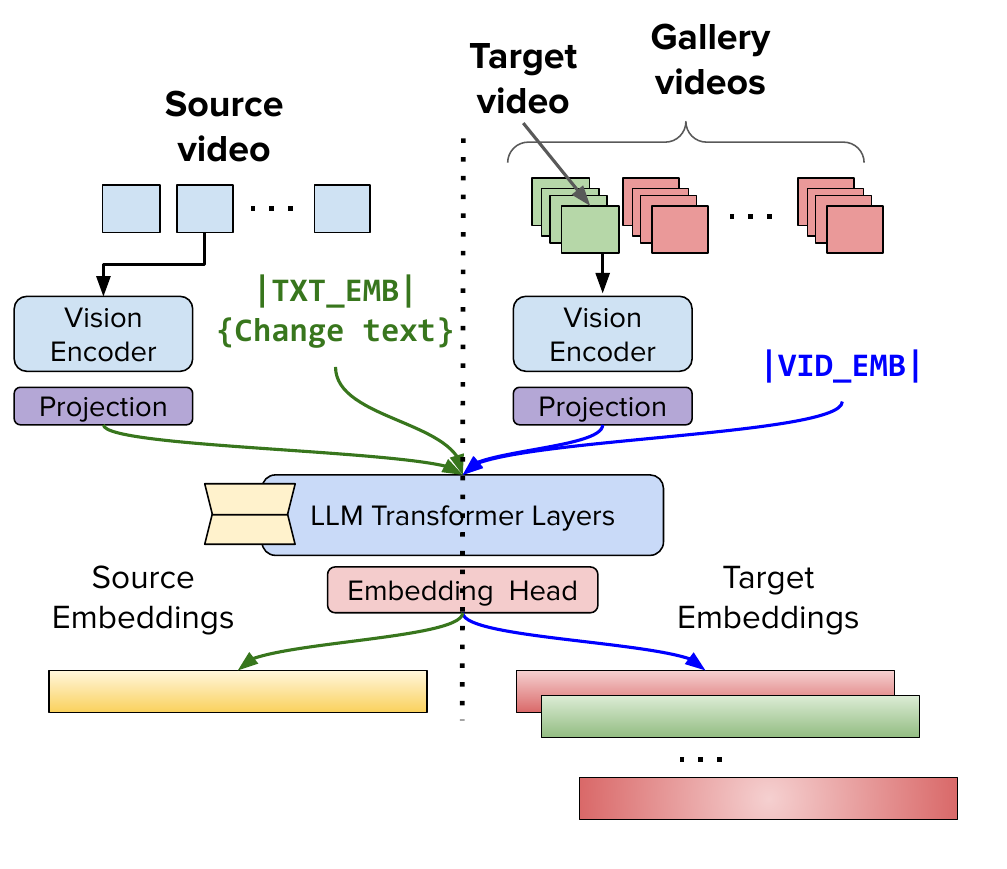}
    \caption{Composed Video Retrieval Inference.}
    \label{fig:covr}
    \vspace{-1em}
\end{figure*}

As illustrated in Fig.~\ref{fig:covr}, inference in the composed-video-retrieval (CoVR) pipeline begins by encoding the source clip with a spatio-temporal vision encoder whose output tokens are linearly projected into the hidden dimension of the downstream large-language-model (LLM) transformer. A single text-edit token, \texttt{|TXT\_EMB|}, embeds the user’s natural-language instruction and is concatenated to the projected visual tokens, enabling early fusion of visual context and linguistic modification. The resulting sequence is processed by several frozen LLM transformer layers, whose final hidden states are pooled to yield a compact query representation that jointly reflects the appearance of the source clip and the requested transformation. In parallel, every clip in the retrieval gallery is independently passed through the same vision-encoder $\rightarrow$  projection $\rightarrow$ LLM steps—without the text token—to obtain target embeddings via a lightweight embedding head. At retrieval time, cosine (or dot-product) similarity is computed between the composed query embedding and all target embeddings, and the gallery video with maximal similarity is returned as the predicted target.

\section{Detailed Overview of Stage 1 Pretraining}
\label{sec:pretraining}

During Stage 1 joint pre-training (Fig.~\ref{fig:pretraining}) the model is optimized with both a causal-language-model (LM) objective and a video–text contrastive objective, each applied in a separate forward pass that nevertheless shares parameters.
First, a batch of videos is fed through a frozen spatio-temporal vision encoder; its output tokens are linearly projected into the hidden dimension of a large language model (LLM) that has been lightweight-fine-tuned via LoRA adapters (modules marked with fire symbol). These visual tokens are prepended to the textual prompt ``<image> … <image> Describe this video.'' and processed by the LLM. Two heads branch from the final hidden states: (i) a language-modeling head generates a caption autoregressively, whose cross-entropy with the ground-truth caption constitutes the language-modeling loss; and (ii) an embedding head pools the hidden states to yield a fixed-length video embedding.

In the second forward pass, only caption strings are supplied—again through the same LLM + LoRA stack, ensuring weight sharing between modalities—and their hidden states are mapped by the embedding head to caption embeddings. The embedding corresponding to the correct caption serves as the positive partner for the video embedding, while captions of the remaining videos in the batch act as negatives. A standard InfoNCE objective then enforces contrastive alignment between the video embedding and its positive caption while repelling mismatched pairs.

The two losses are summed, so the model simultaneously learns to generate fluent video descriptions and to discriminate between semantically consistent versus inconsistent video–text pairs, yielding representations that are effective for both captioning and downstream retrieval tasks.

\section{Detailed Overview of Stage 3: Temporal localization training}
\label{sec:temporaltraining}

As depicted in Fig.~\ref{fig:localization}, Stage 3 (Temporal Localization Task) fine-tunes the model for temporal localization by contrasting text queries with candidate video clips extracted from long-form footage**.  Given an annotated clip-level caption (e.g., ``playing with glow slime in the dark''), a small set of sparsely sampled frames from the entire video is first prepended to the caption prompt to furnish coarse visual context, after which the sequence is routed through the LoRA-adapted LLM and pooled by the embedding head to yield a contextualized text embedding.  In parallel, the full video is partitioned with a fixed-stride sliding window; each windowed segment is passed through the same vision-encoder $\rightarrow$ projection $\rightarrow$ LLM $\rightarrow$ embedding-head stack to obtain video-clip embeddings.  The clip whose temporal bounds coincide with the ground-truth annotation is designated the \emph{positive} match, while all other windowed segments within the batch serve as \emph{hard negatives}.  A contrastive InfoNCE loss is then applied between the text embedding and the set of clip embeddings, driving the model to maximise similarity with the positive segment and to minimise similarity with temporally misaligned negatives.  This procedure equips the shared embedding space with fine-grained temporal sensitivity, enabling the model to accurately localise textual events within extended, untrimmed videos without requiring explicit frame-level supervision.


\section{Intermediate Pre-training Data Balancing}
\label{sec:datadist}

\begin{figure}[h]
    \centering
    \includegraphics[width=\linewidth]{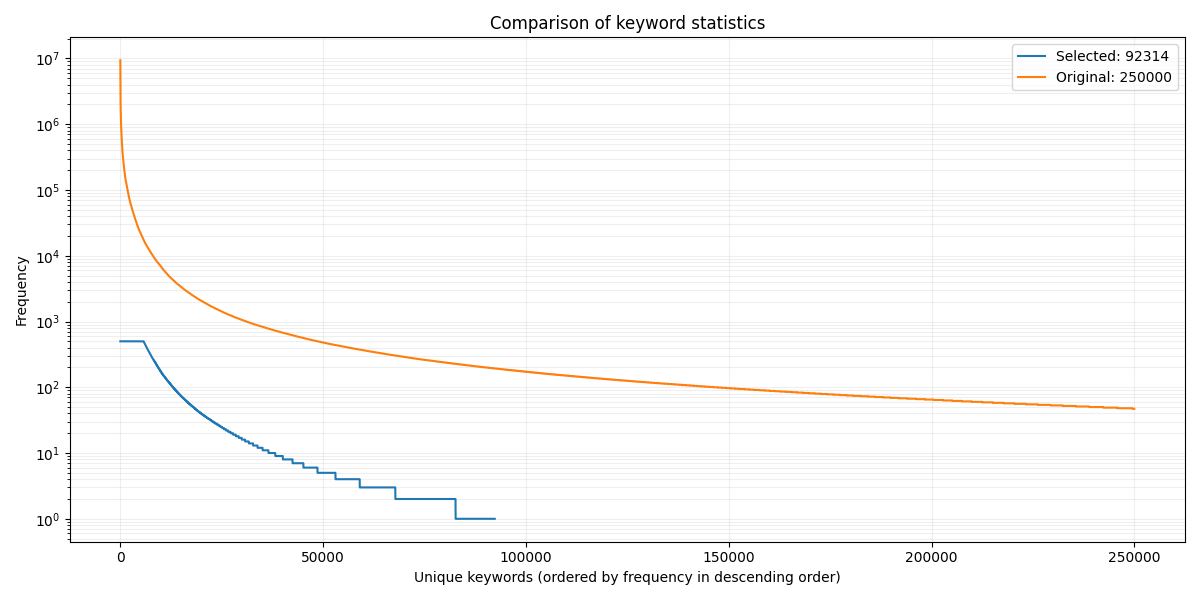}
    \caption{We ensure balance of concepts in our dataset by sub-sampling videos from original dataset by undersampling common concepts. The y-axis is log-scale.}
    \vspace{-0.5em}
    \label{fig:undersampling}
\end{figure}

The Shutterstock dataset is extensively labelled with keywords, which we use to select a balanced set of videos for our intermediate pre-training dataset. Keywords with fewer than 30 occurrences are excluded, and up to 500 videos per remaining keyword are added to the candidate pool, starting with the most frequent keywords, to ensure a balanced and representative selection. The original frequency distribution of keywords, and the final achieved distribution are shown in Fig.~\ref{fig:undersampling}. Note that as each video has multiple keywords, this is not a proper probability distribution.

\section{High Quality Data Statistics}

Our high quality recaptioning of a subset of shutterstock videos significantly increases the level of detail in the captions from having <10 words per caption on average to having 130+ words on average per caption.

\begin{figure}[h]
    \centering
    \includegraphics[width=\linewidth]{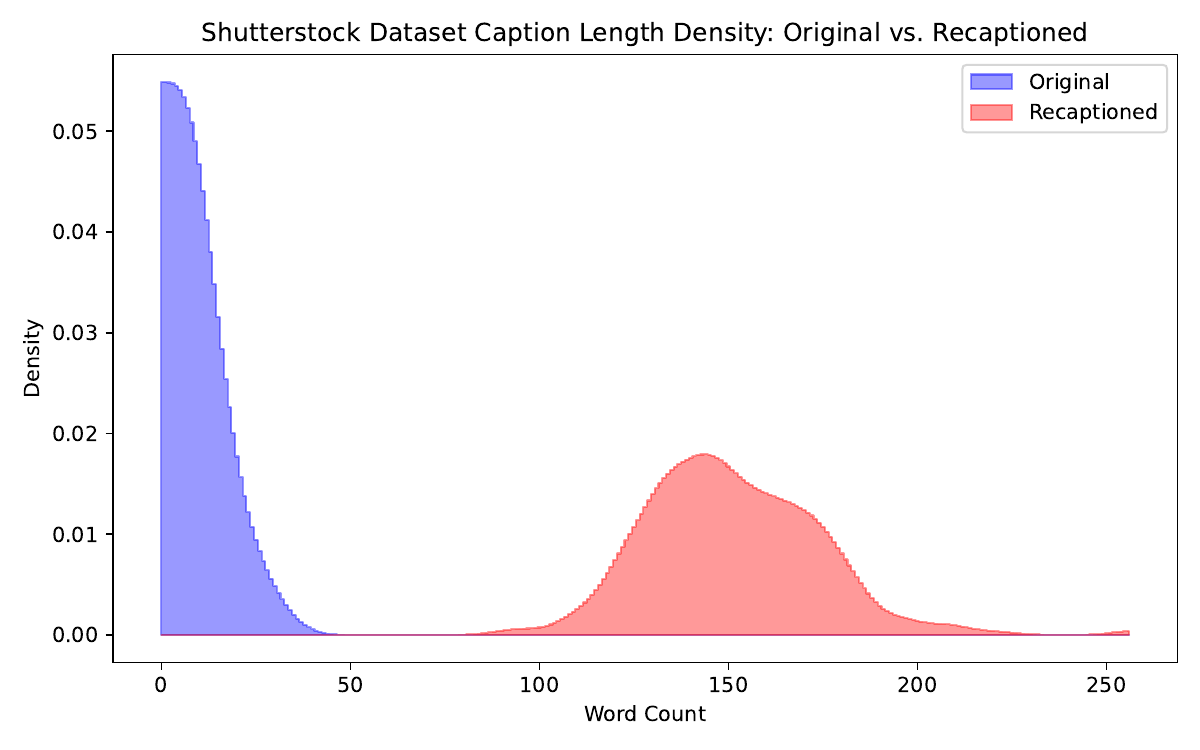}
    \caption{Our captions significantly increases the level of detail. Blue - original captions, Red - Generated.}
    \label{fig:recaptioning}
    
\end{figure}

\begin{figure*}[t]
    \centering
    \includegraphics[width=0.8\linewidth]{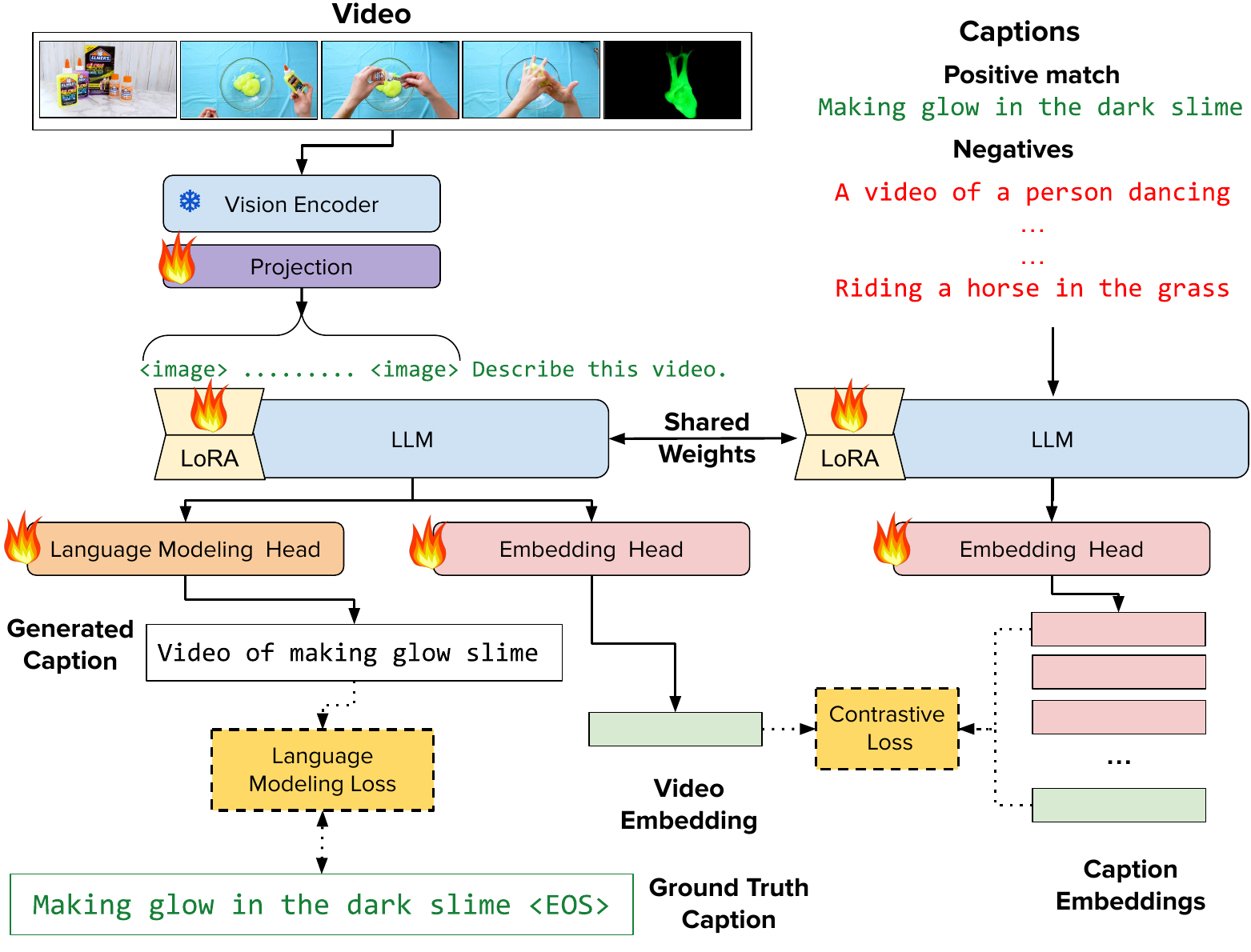}
    \caption{\textbf{Stage 1 Training}. Joint Generative and Contrastive Pre-Training. This stage requires 2 forward passes of the model per step. In the first step, the videos along with prompt are passed to the model and the video caption and video embedding are generated simultaneously. In the second forward pass, only the captions are passed to generate caption embeddings. The ground-truth caption for the video serves as the positive match, whereas the captions for the other videos in the batch are negatives for the contrastive loss.}
    \label{fig:pretraining}
    \vspace{-1em}
\end{figure*}

\begin{figure*}[t]
    \centering
    \includegraphics[width=\linewidth]{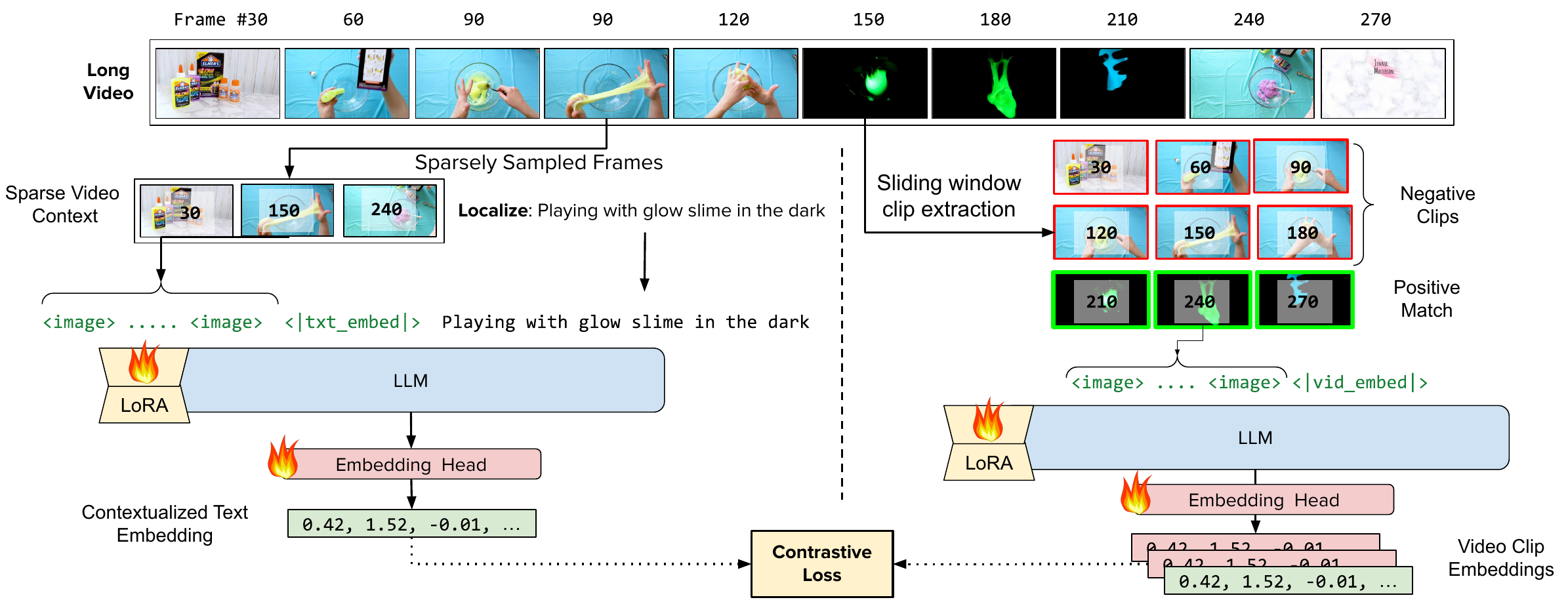}
    \vspace{-1.5em}
    \caption{Stage 3: \textbf{Temporal Localization Task Training}. \underline{Contextualized Text Embeddings} (\textbf{Left Half}): Text embedding are generated based on text input such as ``Playing with glow slime in the dark.'' A selection of sparse frames from the long video is sampled and added to the prompt to provide context. \underline{Sliding Window Video Clip Embeddings} (\textbf{Right Half}): The long video is divided into clips via a sliding window mechanism, which extracts several short clips. Video embeddings for the clips are used to create the positive and hard negative matches for the text, based on the annotations from the dataset. Training is done using the contrastive loss.}
    \label{fig:localization}
     \vspace{-1.5em}
\end{figure*}

\section{Qualitative Video Captioning Results}
\label{sec:qualitative}

\begin{figure*}
    \centering
    \includegraphics[width=\linewidth]{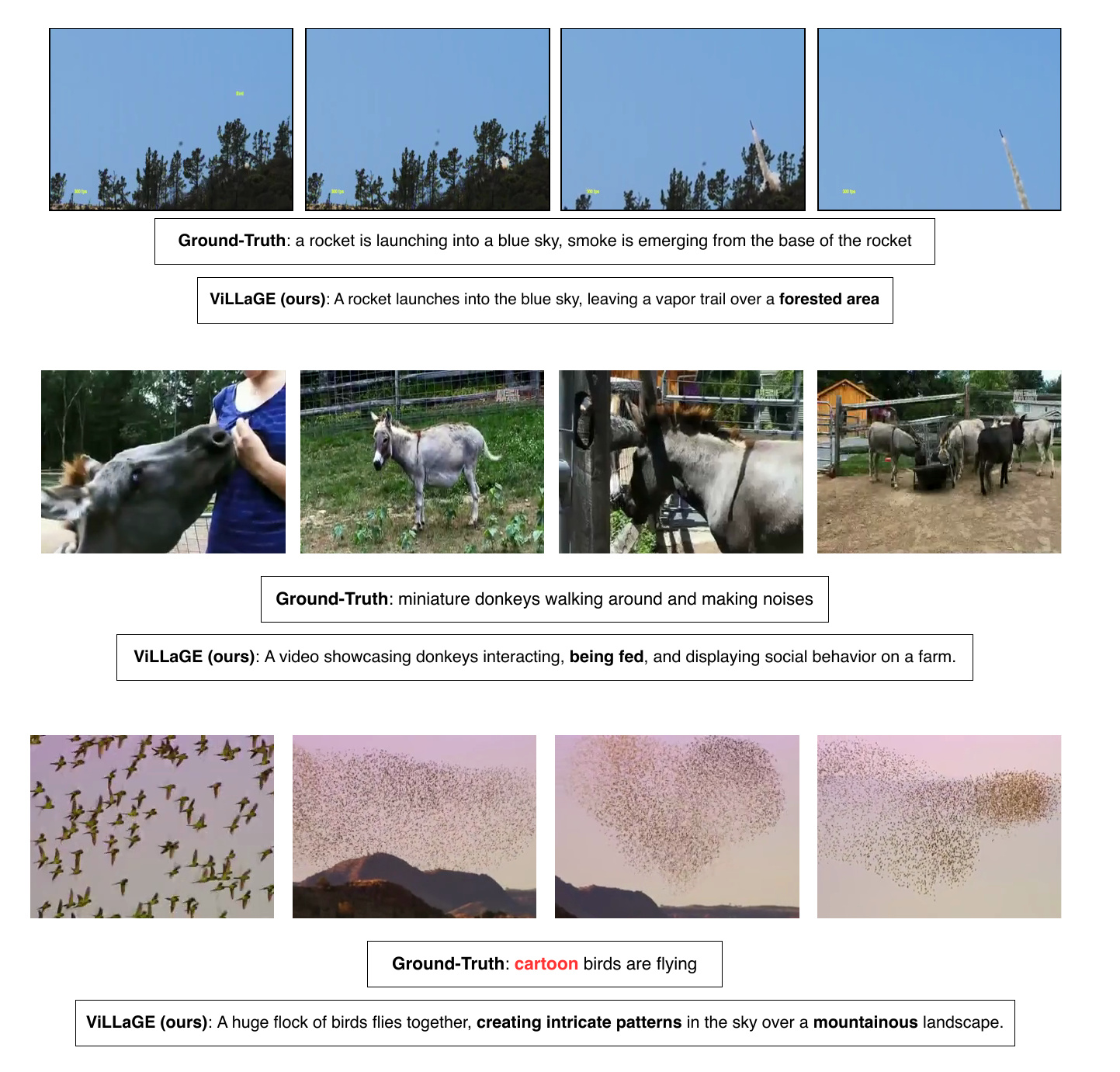}
    \caption{Qualitative Video Captioning results on MSR-VTT}
    \label{fig:qualcaption}
\end{figure*}

ViLL-E is able to generate rich video captions exceeding even the quality of some ground-truth captions in datasets. Some examples of this on the MSR-VTT dataset are provided in Figure~\ref{fig:qualcaption}. The figures compare ground-truth descriptions with those generated by the ViLL-E system, highlighting its ability to enrich scene interpretations. For a rocket launch, the ground-truth describes a simple ascent with smoke at the base, while ViLL-E adds detail about a vapor trail over a forested area. In a depiction of miniature donkeys, the ground-truth focuses on basic movement and noise, whereas ViLL-E elaborates on their interactions, feeding, and social behavior on a farm. Similarly, a scene of cartoon birds flying is expanded by ViLL-E into a vivid portrayal of a large flock creating intricate patterns over a mountainous landscape. These comparisons underscore ViLL-E's capacity to provide more nuanced and context-rich scene descriptions.

\section{Visualization of Video Embeddings}

Text and Video embeddings for video from a held out val set for ShutterStock are visualized after being reduced to two dimensions with Barnes-Hut t-SNE (perplexity = 30, init = PCA), producing two coordinates for every video–caption pair while approximately maintaining local neighbourhood structure. To aid qualitative inspection, we cluster the caption embeddings with K-means (k = 20). The dominant, non-stop-word within each cluster’s captions is used as an interpretable label (e.g., ``beach'', ``office'', ``smiling''). The visualization is rendered as a 4 by 5 grid: each panel isolates one cluster, plotting video embeddings as filled circles and caption embeddings as crosses. Grey lines connect every video point to its paired caption point, revealing intra-cluster fine-grained alignment patterns. The resulting figure offers an intuitive map for understanding the retrieval performance of our model across various domains; a high number of intersecting lines indicates poor performance in that cluster and vice versa. Our model exhibits strong performance on videos with concepts like ``couple'', ``flying'', ``summer'' etc.

    

\begin{figure*}[htbp]
    \centering
    \includegraphics
      [width=\linewidth]
      {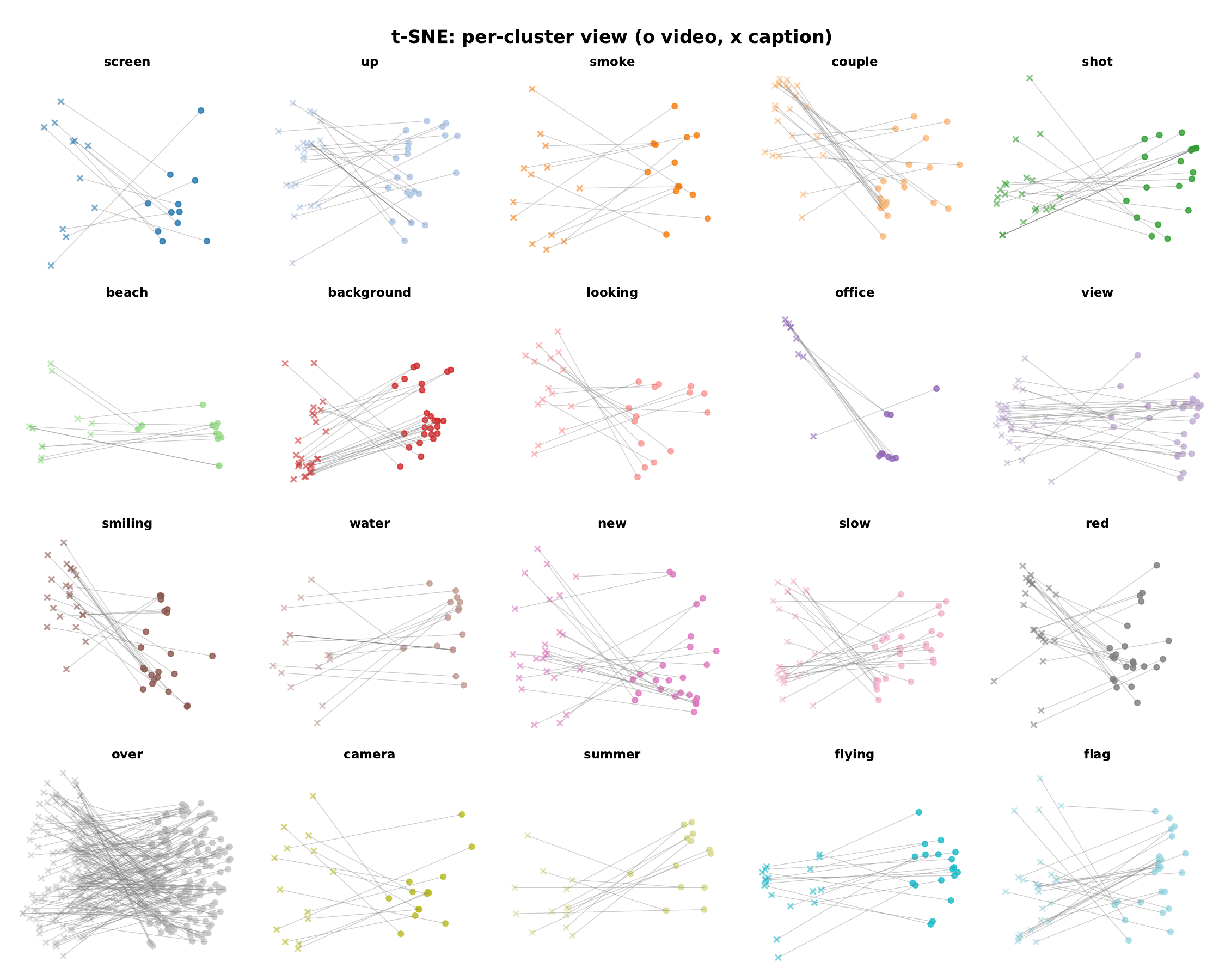}%
    \caption{\parbox{\textwidth}{Illustrating our model's retrieval embeddings and their alignment in various sub-domains of videos}}
    \label{fig:bleed}
\end{figure*}






\end{document}